\documentclass[letterpaper]{article} 
\usepackage{aaai2026}  
\usepackage{times}  
\usepackage{helvet}  
\usepackage{courier}  
\usepackage[hyphens]{url}  
\usepackage{graphicx} 
\usepackage{natbib}  
\usepackage{caption} 
\usepackage{algorithm}
\usepackage{algorithmic}

\usepackage{newfloat}
\usepackage{listings}
\DeclareCaptionStyle{ruled}{labelfont=normalfont,labelsep=colon,strut=off} 
\floatstyle{ruled}
\newfloat{listing}{tb}{lst}{}
\floatname{listing}{Listing}

\usepackage{amsmath}
\usepackage{amssymb}
\usepackage{mathtools}
\usepackage{amsthm}

\usepackage{algorithmic}
\usepackage{algorithm}
\usepackage{booktabs}
\usepackage{multirow}
\usepackage{makecell}
\usepackage{subcaption}

\DeclareMathOperator\artanh{artanh}

\newtheorem{theorem}{Theorem}

\theoremstyle{plain}

\newtheorem{lemma}[theorem]{Lemma}

\theoremstyle{definition}
\newtheorem{definition}[theorem]{Definition}

\theoremstyle{remark}

\title{Hyperbolic Continuous Structural Entropy for Hierarchical Clustering}
\author{
    Guangjie Zeng\textsuperscript{\rm 1}, 
    Hao Peng\textsuperscript{\rm 1}, 
    Angsheng Li\textsuperscript{\rm 1}, 
    Li Sun\textsuperscript{\rm 2}, 
    Chunyang Liu\textsuperscript{\rm 3}, 
    Shengze Li\textsuperscript{\rm 4},
    Yicheng Pan\textsuperscript{\rm 1}\thanks{Corresponding author.}, 
    Philip S. Yu\textsuperscript{\rm 5}
}
\affiliations{
    \textsuperscript{\rm 1}School of Computer Science and Engineering, Beihang University, China\\
    \textsuperscript{\rm 2}School of Control and Computer Engineering, North China Electric Power University, China  \\
    \textsuperscript{\rm 3}Didi Chuxing, China \\
    \textsuperscript{\rm 4}Academy of Military Science, China \\
    \textsuperscript{\rm 5}Department of Computer Science, University of Illinois Chicago, USA \\
    \{zengguangjie, penghao, angsheng, yichengp\}@buaa.edu.cn, ccesunli@ncepu.edu \\
    liuchunyang@didiglobal.com, lsz86591989@163.com, psyu@uic.edu

}

\begin{document}

\maketitle

\begin{abstract}
Hierarchical clustering is a fundamental machine-learning technique for grouping data points into dendrograms.
However, existing hierarchical clustering methods encounter two primary challenges:
1) Most methods specify dendrograms without a global objective.
2) Graph-based methods often neglect the significance of graph structure, optimizing objectives on complete or static predefined graphs.
In this work, we propose \textbf{Hyp}erbolic \textbf{C}ontinuous \textbf{S}tructural \textbf{E}ntropy neural networks, namely HypCSE, for structure-enhanced continuous hierarchical clustering.
Our key idea is to map data points in the hyperbolic space and minimize the relaxed continuous structural entropy (SE) on structure-enhanced graphs. 
Specifically, we encode graph vertices in hyperbolic space using hyperbolic graph neural networks and minimize approximate SE defined on graph embeddings.
To make the SE objective differentiable for optimization, we reformulate it into a function using the lowest common ancestor (LCA) on trees and then relax it into continuous SE (CSE) by the analogy of hyperbolic graph embeddings and partitioning trees.
To ensure a graph structure that effectively captures the hierarchy of data points for CSE calculation, we employ a graph structure learning (GSL) strategy that updates the graph structure during training.
Extensive experiments on seven datasets demonstrate the superior performance of HypCSE.
\end{abstract}

\begin{links}
    \link{Code}{https://github.com/SELGroup/HypCSE}
    \link{Datasets}{https://github.com/SELGroup/HypCSE/datasets}
\end{links}

\section{Introduction}

Hierarchical clustering is a classic unsupervised machine-learning technique that groups data points into nested clusters organized as a cluster tree known as a dendrogram \cite{ran2023comprehensive}.
Unlike partitioning clustering, which divides data points into a predefined number of clusters, hierarchical clustering captures relationships between data points at a finer granularity through nested clusters, without requiring the number of clusters to be specified in advance.
It has broad applications including image analysis \cite{yan2021unsupervised}, bioinformatics \cite{chen2023incorporating}, and medicine \cite{ciaramella2020data}.

Despite their success, conventional hierarchical clustering algorithms specify dendrograms procedurally \cite{chierchia2019ultrametric}.
Among them, agglomerative methods \cite{gower1969minimum} iteratively merge closest cluster pairs, while divisive methods \cite{moseley2017approximation} iteratively split clusters by flat clustering.
Thus, they encounter \textbf{Challenge 1:} most of them lack a global objective and often lead to suboptimal dendrograms.
Recently, Dasgupta's cost \cite{dasgupta2016cost} and its extensions \cite{cohen2019hierarchical,wang2020improved} have been proposed to evaluate dendrogram quality globally.
From an information-theoretic perspective, SE \cite{li2016structural,li2024science} measures dendrogram quality by quantifying the minimum length of random walks on graphs.
These cost functions facilitate the analysis of algorithms and the comparison of dendrograms.

Based on the optimization algorithms for cost functions they adopted, hierarchical clustering methods can be divided into discrete and continuous methods.
Continuous methods \cite{monath2017gradient,monath2019gradient,chierchia2019ultrametric,chami2020trees,zugner2022end} relax a given cost function to be differentiable and perform optimization using gradient-based techniques.
Their chosen cost functions include Dasgupta's cost \cite{dasgupta2016cost} and Tree-Sampling-Divergence \cite{charpentier2019tree}, both of which are defined on graphs.
Compared to discrete methods, continuous methods offer flexibility since they can be integrated into commonly used end-to-end learning pipelines.
However, existing continuous methods optimize objectives on complete graphs \cite{chami2020trees} or predefined graphs \cite{chierchia2019ultrametric,zugner2022end}.
Complete graphs can be cluttered with trivial edges from data noise, while predefined graphs are static and unable to learn from data features during training.
As a result, they encounter \textbf{Challenge 2:} they may not fully utilize data features or capture hierarchies by optimizing objectives on static graphs.

\begin{figure}[t]
\centering
\includegraphics[width=0.95\linewidth]{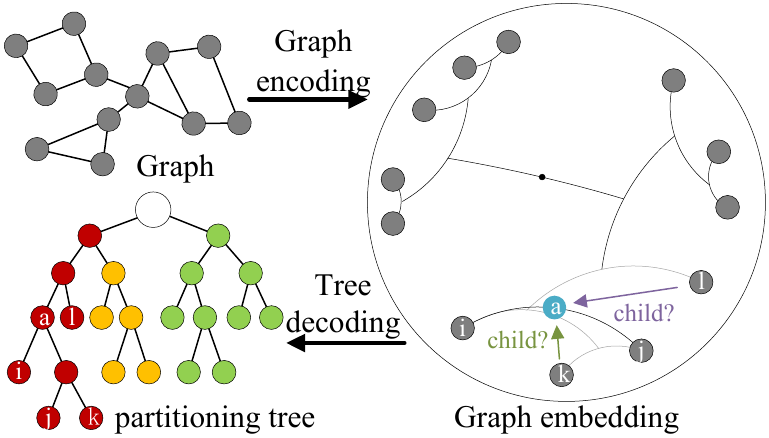}
\caption{HypCSE overview.
Graphs are encoded as hyperbolic embeddings by minimizing CSE.
Partitioning trees are decoded from embeddings for hierarchical clustering.
}
\label{fig_demo}
\end{figure}

To address the aforementioned challenges, we introduce continuous SE (CSE) in hyperbolic space for hierarchical clustering.
CSE is defined on a graph and its hyperbolic space embedding, measuring the quality of the graph embedding in capturing the graph's hierarchy.
We formulate the SE objective as the weighted average of LCAs' volumes of graph vertices.
In hyperbolic space, graph embeddings can be analogous to partitioning trees \cite{chami2020trees}.
We quantify the CSE of graph embedding by approximating the volumes of LCAs on the analogous partitioning tree.
Unlike DSI in LSEnet \cite{sunlsenet}, which relaxes SE by level-wise assignment for flat partitioning clustering, our CSE relaxes SE by analogizing hyperbolic graph embeddings and binary partitioning trees for hierarchical clustering.
Next, we optimize the graph embedding for hierarchical clustering by minimizing the CSE, as depicted in Figure \ref{fig_demo}.
Given a constructed graph from input data, we encode it as a graph embedding in hyperbolic space and optimize the corresponding CSE via gradient descent.
The optimized graph embedding is then decoded as a binary partitioning tree, serving as the hierarchical clustering result.
Furthermore, we design a Hyperbolic CSE Neural Network (HypCSE) for CSE optimization on adaptive graphs updated during training.
To avoid the use of predefined, inappropriate graphs, we employ a GSL strategy to learn more informative graphs.
We construct an anchor graph as the anchor view and learn a new graph as the learner view, applying contrastive learning to guide the graph learner.
The anchor graph is updated according to the learner graph through bootstrapping.
This strategy helps learn anchor graphs with clear hierarchies and high discrimination among classes for CSE optimization.

The main contributions are summarized as follows:
(1) We devise CSE by relaxing discrete SE to measure the quality of graph embeddings in hyperbolic space for capturing hierarchy.
(2) We propose HypCSE via CSE for hierarchical clustering, where a GSL strategy is adopted to learn better graphs for CSE optimization.
(3) Extensive experiments on 7 standard datasets demonstrate the superiority of HypCSE.

\section{Preliminaries}

\subsection{Graph-Based Hierarchical Clustering}
For a dataset $\mathbf{X}$ with $n$ data points, graph $G$ representing pairwise similarities (or dissimilarities) is denoted as $G = (V, E, \mathbf{W})$, where $V = \{ v_1,v_2,\ldots,v_n \}$ is the set of vertices corresponding to data points in $\mathbf{X}$, $E$ is the set of edges, and $\mathbf{W}$ consists of edge weights measuring pairwise similarities (or dissimilarities) between data points.
A hierarchical clustering algorithm generates a dendrogram from $\mathbf{X}$ along with $G$.
This dendrogram is an unweighted rooted tree $\mathcal{T}$ with $n$ leaves corresponding $n$ data points and internal nodes corresponding to nested clusters.
For two leaves $\mathcal{T}_i$ and $\mathcal{T}_j$ in $\mathcal{T}$, their LCA is denoted as $\mathcal{T}_i \vee \mathcal{T}_j$.

\subsection{Hyperbolic Space}
Hyperbolic space is a space with negative curvature that has advantages in modeling hierarchical structure compared to flat Euclidean space \cite{peng2021hyperbolic,sunlsenet,www25RiemannGFM,nips25DeepMPNN}.
Among the several isomorphic models of hyperbolic space, two commonly used ones are the Lorentz model and the Poincar\'e model.
Specifically, a $d$-dimensional Lorentz model $\mathbb{L}^{\kappa,d}$ with curvature $\kappa$ represents a manifold embeded in the $d+1$ dimensional Minkowski space, defined as $\mathbb{L}^{\kappa,d} = \{ \mathbf{x} \in \mathbb{R}^{d+1} :  \langle \mathbf{x},\mathbf{x} \rangle _\mathbb{L} = \frac{1}{\kappa} \}$, where $\langle , \rangle _\mathbb{L}$ is the Minkowski inner product defined as $\langle \mathbf{x},\mathbf{y} \rangle _\mathbb{L} = \mathbf{x}^T \mathbf{R}^\mathbb{L} \mathbf{y}$, $\mathbf{R}^\mathbb{L} \in \mathbb{R}^{(d+1) \times (d+1)}$ is a diagonal matrix with entries of $1$s except for the first one being $-1$.
Lorentz distance between points $\mathbf{x}$ and $\mathbf{y}$ is defined as $d_\mathbb{L}\mathrm{arcosh}(-\langle \mathbf{x},\mathbf{y} \rangle)$.
Lorentz norm of $\mathbf{x}$ is defined as $\| \mathbf{x} \|_\mathbb{L} = \sqrt{\langle \mathbf{x},\mathbf{x} \rangle_\mathbb{L}}$.
A $d$-dimensional Poincar\'e model $\mathbb{B}^{\kappa, d}$ with curvature $\kappa$ is defined as $\mathbb{B}^{\kappa, d} = \{ \mathbf{u} \in \mathbb{R}^d: \| \mathbf{u} \| < \frac{1}{\kappa} \}$, which can be given by projecting each point of Lorentz model $\mathbb{L}^{\kappa,d}$ onto the hyperplane $\mathbf{x}^{0,d} = 0$.
For a point $\mathbf{x} = (x_0, x_1,\ldots,x_d) \in \mathbb{R}^{d+1}$ in the Lorentz model, the transformed point $\mathbf{u}$ in the Poincar\'e model is given by $\mathbf{u} = \frac{(x_1,\ldots,x_d)}{1+x_0}$.
\section{Methodology}

\begin{definition}[Structural Entropy (\cite{li2016structural})]\label{def:se}
Given an undirected weighted graph $G = (V,E,\mathbf{W})$ and an associated rooted tree $\mathcal{T}$, the SE of $G$ on $\mathcal{T}$ is defined as
\begin{equation}\label{eq:se}
    \mathcal{H}^\mathcal{T}(G) = \sum_{\alpha\in\mathcal{T},\alpha\neq\lambda}\mathcal{H}^\mathcal{T}(G;\alpha)
    = \sum_{\alpha\in\mathcal{T},\alpha\neq\lambda} - \frac{g_\alpha}{\mathcal{V}_G}\log_2\frac{\mathcal{V}_\alpha}{\mathcal{V}_{\alpha^-}},
\end{equation}
where $\mathcal{H}^\mathcal{T}(G; \alpha)$ is the SE assigned to an non-root node $\alpha$ in $\mathcal{T}$.
The rooted partitioning tree $\mathcal{T}$ of $G$ forms a hierarchical clustering of data points.
Each tree node $\alpha \in \mathcal{T}$ is associated with a vertex set $T_{\alpha}$, where the root node $\lambda$ is associated with $T_{\lambda} = V$ and each leaf node $\nu$ is associated with $T_{\nu}$ containing only one vertex in $V$.
For each non-leaf node $\alpha \in \mathcal{T}$, the child nodes of $\alpha$ are associated with disjoint vertex subsets, whose union is $T_{\alpha}$.
The parent node of $\alpha$ is denoted as $\alpha^-$.
The notation $g_{\alpha}$ is the cut, i.e., the sum of graph edge weights with exactly one endpoint in $T_{\alpha}$.
Notations $\mathcal{V}_{\alpha}$ and $\mathcal{V}_{G}$ are the volumes, i.e., the sum of node degrees in $T_{\alpha}$ and $T_G$, respectively.
The SE of graph $G$ is the minimum one among all possible partitioning trees, which is defined as
\begin{equation}
    \mathcal{H}(G) = \min_{\mathcal{T}} \mathcal{H}^\mathcal{T}(G), \quad
     \mathcal{T}^*=\arg_{\mathcal{T}}\min  \mathcal{H}^{\mathcal{T}}(G),
\end{equation}
where $\mathcal{T}^*$ is the optimal partitioning tree that best eliminates uncertainty in $G$ by characterizing the hierarchical topology.
\end{definition}

\begin{lemma}[Minimum Structural Entropy (\cite{zhang2021supertad})]\label{lemma:mse}
Given an undirected weighted graph $G$, a binary partitioning tree $\mathcal{T}^*$ of the minimum structural entropy exists.
\end{lemma}
Lemma \ref{lemma:mse} reveals that we can find the minimum SE of $G$ by traversing all possible binary partitioning trees.

\begin{lemma}[Connection to Graph-Based Clustering-Appendix A.1]\label{lemma:conductance}
Given an weighted undirected graph $G=(V, E, \mathbf{W})$, $\rho^{\mathcal{T}}(G) = \mathcal{H}^{\mathcal{T}}(G) / \mathcal{H}^1(G)$ is the normalized SE of $G$ on any possible rooted partitioning tree $\mathcal{T}$, and $\Phi(G) = \min_{S \subseteq V} \frac{\mathrm{cut}(S)}{\min \{ \mathcal{V}_S, \mathcal{V}_{V \backslash S } \}}$ is the graph conductance, the following inequality holds:
\begin{equation}
\rho^\mathcal{T}(G) \geq \Phi(G).
\end{equation}
\end{lemma}
The proof is given in Appendix A. 
We note that graph conductance is a well-defined metric for assessing clustering quality on graphs and has been proven effective for graph-based hierarchical clustering \cite{cheng2006divide}.
Since the one-dimensional SE $\mathcal{H}^1(G)$ is a constant for a given $G$, and it serves as an upper bound for graph conductance (Lemma \ref{lemma:conductance}), minimizing SE in Definition (\ref{def:se}) is likely to also reduce graph conductance.
Furthermore, SE evaluates the quality of the partitioning tree globally, making it a suitable \textit{global} objective for graph-based hierarchical clustering.

\begin{figure*}[t]
\centering
\includegraphics[width=1\linewidth]{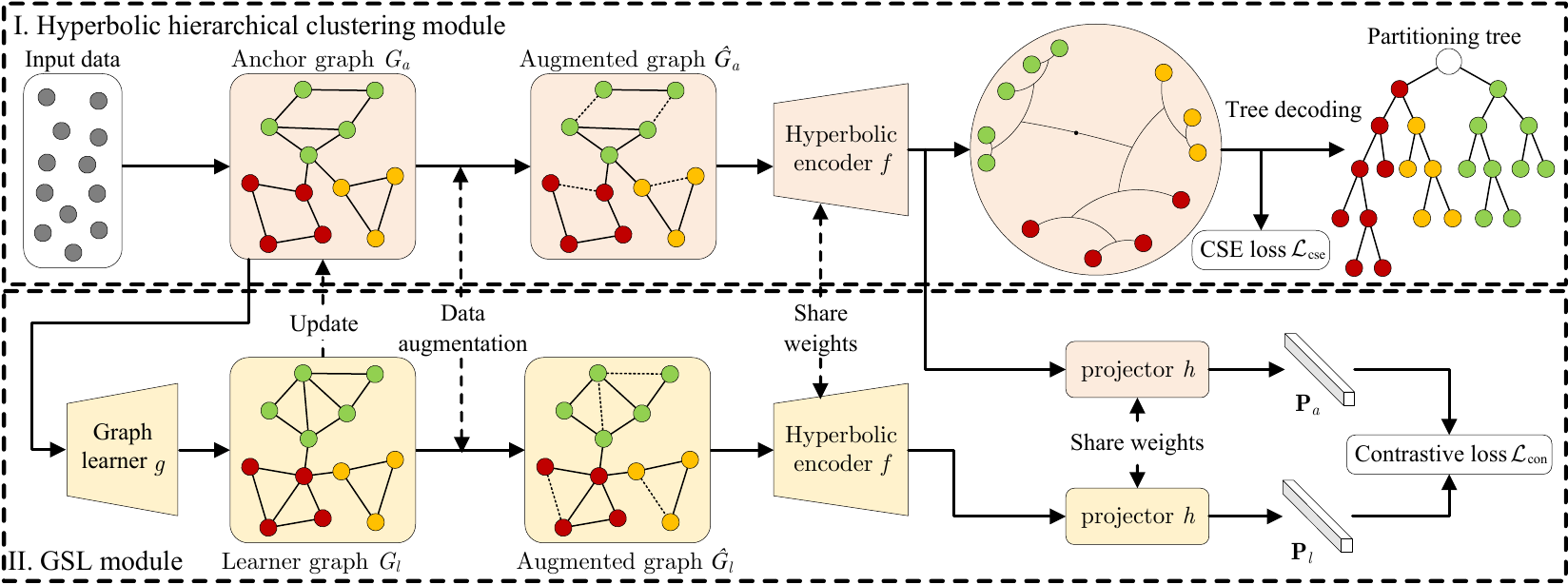}
\caption{Framework of HypCSE.
(I) In the hyperbolic hierarchical clustering module, we construct an anchor graph $G_a$ from the input data, encode it using $f(\cdot)$, and decode it into a binary partitioning tree for hierarchical clustering.
(II) In the GSL module, we learn a leaner graph $G_l$ using graph learner $g(\cdot)$, update $G_a$ from $G_l$, and guide $g(\cdot)$ via contrastive learning.
}
\label{fig_framework}
\end{figure*}

\subsection{Continuous SE in Hyperbolic Space} \label{sec_cse} 
We address \textbf{Challenge 1} by introducing CSE as the global objective in HypCSE.
Both geodesics in hyperbolic space and shortest paths in trees can be viewed as paths with the shortest distances between points.
This relationship allows us to derive hyperbolic LCA analogous to the discrete LCA \cite{chami2020trees}.
To achieve continuous optimization for HypCSE, we reformulate SE based on LCA and derive a differentiable objective CSE by the hyperbolic LCA.

\begin{definition}[Structural Entropy via LCA]\label{def:selca}
Given an undirected weighted graph $G = (V,E,\mathbf{W})$ and an associated rooted tree $\mathcal{T}$, the SE of $G$ on $\mathcal{T}$ can be defined as
\begin{equation}\label{eq:selca}
\begin{aligned}
    \mathcal{H}^\mathcal{T}(G) = \frac{2}{\mathcal{V}(G)} \sum_{(v_i,v_j) \in E} \mathbf{W}_{ij} \log_2 \mathcal{V}_{\mathcal{T}_i \vee \mathcal{T}_j} \\
    - \frac{1}{\mathcal{V}(G)} \sum_{v_i \in V}\mathcal{V}_{\mathcal{T}_i} \log_2 \mathcal{V}_{\mathcal{T}_i},
\end{aligned}
\end{equation}
where $\mathbf{W}_{ij}$ is the edge weight between pair $(v_i,v_j)$, $\mathcal{T}_i$ is the leaf node in $\mathcal{T}$ that contains only one vertex $v_i$, and $\mathcal{V}_{\mathcal{T}_i \vee \mathcal{T}_j}$ is the volume of the LCA of leaf pair $(\mathcal{T}_i,\mathcal{T}_j)$.
For a given graph $G$, the values of the second term in Eq. (\ref{eq:selca}) are the same for all partitioning trees.
\end{definition}

\begin{theorem}[Equivalence (\cite{pan2021information})]\label{theorem:eq}
$\mathcal{H}^\mathcal{T}(G)$ in Def. \ref{def:selca} is equivalent to Eq. (\ref{eq:se}) in Def. (\ref{def:se}).
\end{theorem}


\noindent\textbf{LCA Volume in Hyperbolic Space.}
According to Eq. (\ref{eq:selca}), the calculation of $\mathcal{H}^\mathcal{T}(G)$ requires finding the LCA of leaf node pairs and calculating the volume.
Given the partitioning tree $\mathcal{T}$, the LCA $\mathcal{T}_i \vee \mathcal{T}_j$ of a leaf pair $(\mathcal{T}_i, \mathcal{T}_j)$ is the node on the shortest path between them and closest to the root $\lambda$.
We have $\mathcal{V}_{\mathcal{T}_i \vee \mathcal{T}_j} = \sum_{v_k \in V} \mathcal{V}_{\mathcal{T}_k} \mathbb{I}[\{\mathcal{T}_k|\mathcal{T}_i,\mathcal{T}_j\}]$, where $\mathbb{I}[ \cdot ]$ is the indicator function and $\{\mathcal{T}_k|\mathcal{T}_i,\mathcal{T}_j\}$ means $\mathcal{T}_k$ is a descendant of $\mathcal{T}_i \vee \mathcal{T}_j$.
When leaf nodes $\mathcal{T}_i$ and $\mathcal{T}_j$ (corresponding to graph vertices $v_i,v_j$) are embedded in hyperbolic space, denoted as $\mathbf{z}_i$ and $\mathbf{z}_j$, the shortest path between them is geodesic. 
Their hyperbolic LCA $\mathbf{z}_i \vee \mathbf{z}_j$ can be defined as the point closest to the origin on this geodesic.
The hyperbolic LCA volume is calculated as $\mathcal{V}_{\mathbf{z}_i \vee \mathbf{z}_j} = \sum_{v_k \in V} \mathcal{V}_{\mathbf{z}_k} \mathbb{I}[\{ \mathbf{z}_k | \mathbf{z}_i, \mathbf{z}_j \}]$, where $\{ \mathbf{z}_k | \mathbf{z}_i, \mathbf{z}_j \}$ means $\mathbf{z}_k$ is a descendant of $\mathbf{z}_i \vee \mathbf{z}_j$.

\begin{lemma}
[Descendant via LCA-Appendix A.2]\label{lemma:descendent}
For leaves $\mathcal{T}_i, \mathcal{T}_j$ and $\mathcal{T}_k$ in rooted tree $\mathcal{T}$, $\mathcal{T}_k$ is the descendant of $\mathcal{T}_i \vee \mathcal{T}_j$ if and only if both the following statements hold: 
\begin{equation}
    d_\lambda (\mathcal{T}_i \vee \mathcal{T}_j) \leq d_\lambda (\mathcal{T}_i \vee \mathcal{T}_k), \quad d_\lambda (\mathcal{T}_i \vee \mathcal{T}_j) \leq d_\lambda (\mathcal{T}_j \vee \mathcal{T}_k),
\end{equation}
where $d_\lambda (\mathcal{T}_i)$ is the distance between $\mathcal{T}_i$ and the root $\lambda$ in $\mathcal{T}$.
\end{lemma}

\begin{lemma}
[Distance of geodesic to origin \cite{chami2020trees}
]
\label{lemma:distance}
For two points $(\mathbf{x},\mathbf{y}) \in \mathbb{B}^{1,d}$ where $\mathbf{x} \vee \mathbf{y}$ is the point on the geodesic connecting $\mathbf{x}$ and $\mathbf{y}$ that minimizes the distance to the origin $\mathbf{o}$, let $\mathbf{r}$ be the symmetric point to $\mathbf{x}$ with respect to the circle at infinity, and $\mathbf{o}_{\mathrm{ref}}$ be the symmetric point to $\mathbf{o}$ with respect to the geodesic. We have:
\begin{equation}
\begin{split}
\ d_o^\mathbb{B}(\mathbf{x}\vee \mathbf{y}) &= \frac{\artanh(\| \mathbf{o}_{\mathrm{ref}} \|^2_2)}{2}, \\
\mathrm{where}\ \ \mathbf{o}_{\mathrm{ref}} &= \frac{\| \mathbf{r} \|^2_2 - 1}{\| \mathbf{o}_{\mathrm{invref}} - \mathbf{r} \|^2_2} \cdot (\mathbf{o}_{\mathrm{invref}} - \mathbf{r}) + \mathbf{r}, \\
\mathbf{o}_{\mathrm{invref}} &= \frac{2 \mathbf{x}^\top \cdot \mathbf{y}_{\mathrm{inv}}}{\| \mathbf{x} \|^2_2} \cdot \mathbf{x} - \mathbf{x}, \ \  \mathbf{r} = \frac{\mathbf{x}}{\| \mathbf{x} \|^2_2},
 \\
\mathrm{and} \ \  
\mathbf{y}_{\mathrm{inv}} &= \frac{\| \mathbf{r} \|^2_2 - 1}{\| \mathbf{y} - \mathbf{r} \|^2_2 } \cdot (\mathbf{y} - \mathbf{r}) + \mathbf{r}.
\end{split}
\end{equation}
\end{lemma}

\noindent\textbf{CSE objective function.}
Given a graph $G$ along with the partitioning tree $\mathcal{T}$, minimizing the SE is equivalent to minimizing the following cost function:
\begin{equation} \label{eq:senondiff}
    C^\mathcal{T}_{G} = \sum_{i\sim j} \mathbf{W}_{ij} \log_2 \bigg( \mathcal{V}_{\mathcal{T}_i} + \mathcal{V}_{\mathcal{T}_j} + \sum_{k \in V}^{k \neq i,j} \mathcal{V}_{\mathcal{T}_k} \mathbb{I}[\{\mathcal{T}_k|\mathcal{T}_i,\mathcal{T}_j\}]\bigg),
\end{equation}
where the notation $i\sim j$ denotes the set of edges $(v_i, v_j) \in E$.
Term $\mathbb{I}[\{\mathcal{T}_k|\mathcal{T}_i,\mathcal{T}_j\}]$ is the non-differential term in Eq. (\ref{eq:senondiff}), which is an indicator for whether $\mathcal{T}_k$ is a descendant of $\mathcal{T}_i \vee \mathcal{T}_j$.
According to Lemma \ref{lemma:descendent}, this indicator can be determined by the relationship among hyperbolic distances $d_\lambda (\mathcal{T}_i \vee \mathcal{T}_j)$, $d_\lambda (\mathcal{T}_i \vee \mathcal{T}_k)$, and $d_\lambda (\mathcal{T}_j \vee \mathcal{T}_k)$.
The indicator outputs true when only $d_\lambda (\mathcal{T}_i \vee \mathcal{T}_j)$ is the smallest one.
When leaves $\mathbf{Z}_\mathbb{B} = \{ \mathbf{z}_1,...,\mathbf{z}_n \}$ in $\mathcal{T}$ are embedded in the hyperbolic space of Poincar\'e model, the distance of LCA to the root $d_\lambda (\mathcal{T}_i \vee \mathcal{T}_j)$ can be approximated as the distance of geodesic to the origin $d_o^\mathbb{B} (\mathbf{z}_i \vee \mathbf{z}_j)$, and the indicator can be approximated by a softmax function.
CSE objective is then:
\begin{equation}\label{eq:hypcse}
\mathcal{L}_{\mathrm{cse}}^{\mathbf{Z}_\mathbb{B}}(G) = \sum_{(v_i,v_j) \in E} \mathbf{W}_{ij} \log_2 ( \mathcal{V}_{\mathbf{z}_i} + \mathcal{V}_{\mathbf{z}_j} + \widehat{\mathcal{V}}_{\mathbf{z}_i \vee \mathbf{z}_j} ), 
\end{equation}
\begin{equation}
\widehat{\mathcal{V}}_{\mathbf{z}_i \vee \mathbf{z}_j} =  \sum_{k \in V, k \neq i.j} (\mathcal{V}_{\mathbf{z}_k},0,0) \cdot \sigma_{t_1} \bigg(s^o_{ij}, s^o_{ik}, s^o_{jk}\bigg)^\top,
\end{equation}
where $\sigma_{t_1} (\cdot)$ is the scaled softmax function: $\sigma_{t_1} (\mathbf{x})_i = e^{\mathbf{x}_i/t_1} / \sum_j e^{\mathbf{x}_j / t_1}$, $s^o_{ij} = r_1 - d_o^\mathbb{B}(\mathbf{z}_i \vee \mathbf{z}_j)$, and $t_1$ and $r_1$ are hyperparameters.


\subsection{Hyperbolic Hierarchical Clustering}\label{sec:hyphc}
Hereafter, we elaborate on achieving hierarchical clustering in hyperbolic space by minimizing CSE objective $\mathcal{L}_{\mathrm{cse}}^\mathbf{Z}$.
As depicted in Figure \ref{fig_framework} (I), the hyperbolic hierarchical clustering module consists of three steps: graph construction, hyperbolic encoding, and partitioning tree decoding.
Given that CSE is defined on graphs, for input data $\mathbf{X}$ containing $n$ data points, we construct a similarity graph $G$ and sparsify it to eliminate noise in data.
First, we construct a weighted undirected graph $G = (V, E, \mathbf{W}, \mathbf{X})$ from $\mathbf{X}$.
In practice, we measure the similarity between data points by Gaussian kernel with kernel width $\sigma=1$ as edge weights and sparsify $G$ by retaining the top $k=10$ edges of each vertex.
Next, we encode $G$ via the Lorentz Convolution $\mathrm{LConv}$ \cite{chen2021fully} into an embedding $\mathbf{Z}$ in hyperbolic space, and minimize $\mathcal{L}_{\mathrm{cse}}^\mathbf{Z}$ via gradient descent optimization.
This embedding $\mathbf{Z}$ has a close correspondence to tree metrics \cite{sarkar2011low}, where each vertex embedding $\mathbf{z}_i$ corresponds to a tree leaf and LCA of leaves form the partitioning tree \cite{chami2020trees}.
Finally, we heuristically decode $\mathbf{Z}$ into a binary partitioning tree for hierarchical clustering.

\noindent\textbf{Hyperbolic graph embedding.} \label{sec_graph_embedding}
Given a constructed attributed graph $G = (V, E, \mathbf{W}, \mathbf{X}) $, we encode it into a hyperbolic embedding $\mathbf{Z} \in \mathbb{L}^{\kappa,d}$ via $\mathrm{LConv}$ \cite{chen2021fully}.
Two key components of $\mathrm{LConv}$ are the Lorentz linear layer $\mathrm{LLinear}$ and the Lorentz attention-based aggregation layer $\mathrm{LAgg}$.
For a graph vertex $v_i$ with feature $\mathbf{x}_i$, its linear transform is defined as follows:
\begin{equation}\label{eq_llinear}
\mathrm{LLinear}(\mathbf{x}_i) = 
\begin{bmatrix}
\sqrt{\| \phi(\Theta \mathbf{x}_i, \mathbf{b}) \|^2 - \frac{1}{\kappa}} \\ 
\phi(\Theta\mathbf{x}_i, \mathbf{b})
\end{bmatrix},
\end{equation}
where $\Theta$ and $\mathbf{b}$ are parameters, and $\phi$ is an operation function.
The aggregation of $\mathbf{x}_i$ is defined as follows:
\begin{equation}
\begin{aligned}
w_{ij} &= \frac{\exp (-\frac{1}{\sqrt{ \mathrm{dim}(\mathbf{x}_i) }} d_{\mathbb{L}}^2 (\mathbf{q}_i, \mathbf{k}_j) )}{ \sum_{l=1}^n   \exp (-\frac{1}{\sqrt{ \mathrm{dim}(\mathbf{x}_i) }} d_{\mathbb{L}}^2 (\mathbf{q}_i, \mathbf{k}_l) ) }, \\
\mathrm{LAgg}(\mathbf{x}_i) &= \frac{\sum_{v_j \in \mathcal{N}(v_i) } w_{ij} \mathbf{v}_j }{ \sqrt{-\kappa} | \|  \sum_{v_l \in \mathcal{N}(v_i) } w_{il} \mathbf{v}_l  \|_{\mathbb{L}} | },
\end{aligned}
\end{equation}
where $\mathrm{dim}(\mathbf{x}_i)$ is the dimension of $\mathbf{x}_i$, $\mathcal{N}(v_i)$ is the neighborhood of vertex $v_i$ in $G$, $\mathbf{q}_i$, $\mathbf{k}_i$ and $\mathbf{v}_i$ are row vectors in the query set $\mathbf{Q}$, key set $\mathbf{K}$, and value set $\mathbf{V}$, respectively.
The Lorentz convolution layer is defined as $\mathrm{LConv} (\mathbf{X} | G) = \mathrm{LAgg}(\mathrm{LLinear}(\mathbf{X})) \in \mathbb{L}^{\kappa, d}$.
In our Hyperbolic encoder $f(\cdot)$, we stack 3 layers of $\mathrm{LConv}$ to obtain the Lorentz embedding $\mathbf{Z}_{\mathbb{L}}$ of G.
Afterwards, we transform $\mathbf{Z}_{\mathbb{L}}$ into Poincar\'e embedding $\mathbf{Z}_{\mathbb{B}}$ to facilitate $\mathcal{L}^\mathbf{Z}_{\mathrm{cse}}$ minimization.

\noindent\textbf{Hyperbolic tree decoding.} \label{sec_tree_decoding}
To minimize SE, conventional algorithms use discrete optimization \cite{li2016structural,pan2021information} to output discrete encoding trees that best characterize the uncertainty of the hierarchical topology of graphs.
HypCSE, however, minimizes the approximate CSE of graphs by optimizing $\mathcal{L}_{\mathrm{cse}}$ on hyperbolic graph embeddings rather than explicit trees.
For the Poincar\'e embedding $\mathbf{Z}_{\mathbb{B}} = \{ \mathbf{z}_1,...,\mathbf{z}_n \}$ encoded by $\mathrm{LConv}$, we find a binary tree by a Single Linkage clustering-like \cite{gower1969minimum} decoding algorithm (Algorithm 1 in Appendix B).
We set each data point $\mathbf{z}_i$ as a cluster (tree leaves) and iteratively merge the two closest clusters $C_\alpha$ and $C_\beta$ into a new cluster.
The closeness of clusters is defined as:
\begin{equation}
    \mathrm{closeness}(C_\alpha, C_\beta) = \min_{\mathbf{z}_i \in C_\alpha,\mathbf{z}_j \in C_\beta}(d_o(\mathbf{z}_i \vee \mathbf{z}_j)).
\end{equation}
To help tree leaves separate in the Poincar\'e disk, we normalize their embeddings to lie on the hyperbolic diameter \cite{chami2020trees}.
We merge clusters until all clusters are merged into one binary partitioning tree.
With K-nearest neighbor graph construction and minimum spanning tree trick (Appendix C.2), the time complexity of tree decoding algorithm can be reduced to $O(n \log n)$.


\subsection{Graph Structure Learning}
Minimizing SE on the constructed graph $G$ guides the hyperbolic encoder $f(\cdot)$ in embedding $G$.
However, we still encounter \textbf{Challenge 2:} optimization on the heuristically constructed static graph $G$ neglects the significance of graph structure and overlooks the information in the original input data.
To address this issue, we adopt a GSL technique \cite{liu2022towards} guided by contrastive learning in hyperbolic space to learn a better graph structure for SE minimization.
As depicted in Figure \ref{fig_framework} (II), the GSL module consists of two steps: graph learning and contrastive learning, as follows.

\noindent\textbf{Graph Learning.}
We take the constructed graph $G = (V, E, \mathbf{X})$ as the anchor graph $G_a = (V, E_a, \mathbf{X})$, which provides stable guidance for GSL.
To learn graph structure from $\mathbf{X}$, we generate a learner graph $G_l$ via a GSL encoder.
First, we apply a graph learner $g(\cdot)$ to learn vertex embeddings $\mathbf{E} = g(\mathbf{X}, E_a)$ for learner graph construction.
Next, we construct the affinity matrix $\mathbf{A}_l$ of $G_l$ by calculating the similarities between pairs of vertex embeddings.
We then select $p$ edges with the highest weights connected to each vertex to form the learner graph edge set $E_l$, resulting in the learner graph $G_l = \{ V, E_l, \mathbf{X} \}$.
To facilitate a more informative anchor graph $G_a$, we update its edge set $E_a$ using $E_l$ as:
\begin{equation}
E_a \leftarrow S_{\tau}(E_a) + S_{1 - \tau}(E_l),
\end{equation}
where $\tau \in (0,1]$ is the decay rate, and $S_{\tau}(\cdot)$ represents a random sampling operator with rate $\tau$.
We choose a large $\tau$ and update $G_a$ after each epoch, allowing it to gradually assimilate new and eliminate erroneous information.
Graph $G_a$ is then used to calculate $\mathcal{L}_{\mathrm{cse}}^\mathbf{Z}(G_a)$ in Eq. (\ref{eq:hypcse}).

\noindent\textbf{Contrastive Learning.}
We adopt contrastive learning to guide the graph learner $g(\cdot)$ and learn more discriminative vertex features.
After obtaining the anchor graph $G_a$ and the learner graph $G_l$, we perform data augmentation by randomly removing edges and masking vertex features.
These graphs are then embedded in hyperbolic space as $\mathbf{Z}_a$ and $\mathbf{Z}_l$ by the hyperbolic encoder $f(\cdot)$.
These embeddings are passed to a projector $h(\cdot)$ with 2 layers of $\mathrm{LLinear}$:
\begin{equation}
\mathbf{P}_a = h(\mathbf{Z}_a), \ \mathbf{P}_l = h(\mathbf{Z}_l).
\end{equation}
Contrastive learning aims to learn representations that can differentiate between similar and dissimilar data points. 
It usually builds upon similarities between data point representation pairs \cite{chen2020simple}.
Unlike representations in Euclidean space, the similarities between vertex features in this context are difficult to define.
Motivated by \cite{ge2023hyperbolic}, we instead minimize the hyperbolic distance of two views from the same vertex and maximize the distance from different vertices.
For two representations $\mathbf{P}_a = \{ \mathbf{p}_a^1,\ldots,\mathbf{p}_a^n \}$ and $\mathbf{P}_l = \{  \mathbf{p}_l^1,\ldots,\mathbf{p}_l^n \}$, the hyperbolic contrastive loss is calculated as follows:
\begin{equation}
\begin{aligned}
\mathcal{L}_{\mathrm{con}} &= \frac{1}{2n} \sum_{i=1}^{n} \big[L(\mathbf{p}_l^i, \mathbf{p}_a^i) + L(\mathbf{p}_a^i, \mathbf{p}_l^i) \big], \\
L(\mathbf{p}_l^i, \mathbf{p}_a^i) &= - \log \frac{\exp \big\{ [r_2 - d_\mathbb{L}(\mathbf{p}_l^i, \mathbf{p}_a^i)] / t_2\big\}}{\sum_{k=1}^n \exp\big\{[r_2 - d_\mathbb{L}(\mathbf{p}_l^i, \mathbf{p}_a^k) ] / t_2\big\} },
\end{aligned}
\end{equation}
where $r_2$ and $t_2$ are hyperparameters.

\subsection{Overall Framework}
We apply several scalability strategies to improve HypCSE's efficiency, reducing the time complexity to $O(n \log n)$ (Appendix C).
To generate more balanced partitioning trees, we ensure that the learned leaf embeddings in $\mathbf{Z}_{\mathbb{L}}$ are scattered around the origin by minimizing the distance between the centroid of leaves and the origin.
We introduce the centroid loss $\mathcal{L}_{\mathrm{cen}}$ based on Lorentz distance \cite{law2019lorentzian} as:
\begin{equation}
    \mathcal{L}_{\mathrm{cen}} = d_o^\mathbb{L} \bigg( \frac{\sum_{j=1}^n\mathbf{z}_j}{\sqrt{-\kappa}| \| \sum_{i=1}^n \mathbf{z}_i \|_\mathbb{L} |} \bigg),
\end{equation}
where $d_o^\mathbb{L}(\mathbf{z}_i)$ is the distance $\mathbf{z}_i$ to origin in Lorentz model.
In summary, the overall objective is formulated as:
\begin{equation}
    \mathcal{L}_{\mathrm{HypCSE}} = \mathcal{L}_{\mathrm{cse}} + \eta_1 \mathcal{L}_{\mathrm{con}} + \eta_2 \mathcal{L}_{\mathrm{cen}},
\end{equation}
where the hyperparameters $\eta_1$ and $\eta_2$ are simply set as 1. 
\section{Experiments}

\begin{table*}
  \centering

  \resizebox{\linewidth}{!}{
  \begin{tabular}{l|cc|cc|cc|cc|cc|cc|cc}
    \toprule
    \multirow{2}{*}{Methods}
    & \multicolumn{2}{c|}{Zoo} & \multicolumn{2}{c|}{Iris} & \multicolumn{2}{c|}{Wine} & \multicolumn{2}{c|}{Br. Cancer} & \multicolumn{2}{c|}{OpticalDigits} & \multicolumn{2}{c|}{Spambase} & \multicolumn{2}{c}{PenDigits}
    \\
     & DP$\uparrow$ & SE$\downarrow$ & DP$\uparrow$ & SE$\downarrow$ & DP$\uparrow$ & SE$\downarrow$ & DP$\uparrow$ & SE$\downarrow$ & DP$\uparrow$ & SE$\downarrow$  & DP$\uparrow$ & SE$\downarrow$ & DP$\uparrow$ & SE$\downarrow$  \\
    \midrule 
    SingleLinkage & \underline{97.6} & 2.037 & 81.2 & 3.483 & 67.9 & 3.909 & 85.1 & 4.977 & 73.3 & 2.839 & 58.9 & 7.180 & 70.0 & 6.125 \\
    BKM & 64.2 & 2.179 & 82.4 & 3.939 & 86.1 & 3.698 & 95.7 & 5.057 & 50.8 & 3.364 & 65.6 & 6.898 & 69.1 & \underline{5.135}  \\
    HDBSCAN & 96.4 & 2.357 & 76.6 & 4.161 & 53.5 & 4.680 & 83.3 & 5.617 & 58.5 & 3.710 & 57.8 & 8.011 & 64.1 & 7.478  \\
    HCSE & 97.3 & \underline{1.929} & \underline{89.7} & 3.593 & 71.1 & 3.819 & 94.2 & 4.319 & 81.5 & 3.011 & 55.2 & 6.599 & \underline{76.9} & 6.877 \\
    SpecWRSC & 95.4 & 2.228 & 83.2 & 3.172 & \underline{93.5} & 3.441 & 95.1 & 4.512 & \underline{85.9} & 2.900 & 55.1 & 8.817 & 65.3 & 7.190 \\
    DPClusterHSBM & 93.6 & 2.264 & 82.9 & \underline{2.997} & 89.5 & \underline{3.404} & 92.9 & \underline{4.169} & 81.0 & \underline{2.517} & 61.0 & \textbf{5.194} & \multicolumn{2}{c}{not converge}\\
    \midrule
    UFit & 93.3 & 2.496 & 81.5 & 3.236 & 78.9 & 3.670 & 95.0 & 4.318 & 69.7 & 3.186 & 59.9 & 6.737 & 70.0 & 6.386 \\
    HypHC & 96.8 & 2.010 & 76.0 & 3.485 & 88.7 & 3.692 & \underline{96.5} & 5.549 & 33.5 & 4.468 & \underline{75.4} & 8.895 & 11.7 & 10.69 \\
    FPH & 89.9 & 2.227 & 85.3 & 3.806 & 92.8 & \textbf{3.102} & 92.6 & \textbf{3.581} & 81.0 & 5.707 & 54.8 & 7.660 & 69.6 & 8.192 \\ 
    HypCSE & \textbf{97.9} & \textbf{1.822} & \textbf{95.1} & \textbf{2.957} & \textbf{93.9} & 3.496 & \textbf{96.8} & 4.342 & \textbf{86.4} & \textbf{2.336} & \textbf{75.5} & \underline{6.668} & \textbf{81.4} & \textbf{4.704} \\
    \bottomrule
    \end{tabular}
    }
    \caption{Hierarchical clustering quality measured in DP ($\%$) and SE. Bold: the best and underline: the runner-up performance.}
    \label{table_main}
\end{table*}

\begin{table}[t]
\centering

 {\fontsize{9}{11}\selectfont
 {\setlength{\tabcolsep}{1.5mm}
\begin{tabular}{ccc|ccccccc}
\toprule
Base & GL & CL                       & Zoo & Iris & Wine & Br. & Opt. & Spam.    \\ \midrule
\checkmark & &                       & 97.8 & 89.7 & 88.8 & 96.0 & 86.2 & 70.1    \\
\checkmark & \checkmark &            & 97.8 & 89.7 & 88.8 & 96.5 & 86.3 & 72.1   \\
\checkmark & &  \checkmark           & 97.7 & 94.1 & 93.4 & 96.6 & 85.9 & 75.4   \\
\checkmark & \checkmark & \checkmark & \textbf{97.9} & \textbf{95.1} & \textbf{93.9} & \textbf{96.8} & \textbf{86.4} & \textbf{75.5}   \\
\bottomrule
\end{tabular}
}
}
\caption{Experimental results (DP $\%$) of ablation study.}
\label{tab:ablation}
\end{table}

\subsection{Experimental Setup}
We adopt two metrics, including Dendrogram Purity (DP) \cite{heller2005bayesian,chami2020trees} and Structural Entropy (SE) \cite{li2016structural}, for hierarchical clustering performance evaluation.
DP is a holistic measure of partitioning trees, defined as the average purity score of LCAs of leaf pairs with the same ground truth labels.
SE of a partitioning tree and its corresponding graph quantifies the amount of uncertainty remaining in this graph, where a lower SE indicates a higher quality of the tree, eliminating more uncertainty in this graph.
We calculate the SEs of the partitioning trees generated by all algorithms in the constructed anchor graphs without performing updates.
In HypCSE, the hyperbolic encoder and projector are based on the Lorentz model with curvature $\kappa=-1$ and optimized via Riemannian Adam \cite{becigneulriemannian} in Geoopt \cite{kochurov2020geoopt}, while the GSL encoder is in Euclidean space and optimized via Adam \cite{kingma2014adam}.
We consistently set $t_1=1000$, $r_1=2$, $t_2=1$, and $r_2=0$, across all datasets, with the sole exception of PenDigits, for which the temperature hyperparameters are $t_1=1$ and $t_2=1000$.
The default value for $\tau$ is $0.9999$.
Further implementation details are in Appendix E.
We conduct all experiments 5 times and report the mean values.

\noindent\textbf{Datasets and Baselines.}
We conduct experiments on 7 clustering datasets from the UCI Machine Learning Repository \cite{UCI}, whose size ranges from 101 to 10,992.
We compare HypCSE against 6 discrete hierarchical clustering methods, including SingleLinkage \cite{gower1969minimum}, BKM \cite{moseley2017approximation}, HDBSCAN\cite{mcinnes2017hdbscan}, SE-based Hierarchical Clustering (HCSE) \cite{pan2021information}, SpecWRSC \cite{laenen2023nearly}, and DPClusterHSBM \cite{imola2023differentially}.
We also compare HypCSE against 3 continuous methods, including UFit \cite{chierchia2019ultrametric}, HypHC \cite{chami2020trees}, and FPH \cite{zugner2022end}.
We adopt the same graphs as the proposed HypCSE for HCSE, SpecWRSC, and FPH since they are defined on graphs and assume the graphs are given.
We adopt the graph construction methods in the original papers for other graph-based methods, i.e., DPClusterHSBM, UFit, and HypHC.

\subsection{Hierarchical Clustering Quality}

In Table \ref{table_main}, we report the performance of discrete and continuous methods for hierarchical clustering across 7 real-world datasets.
HypCSE outperforms its discrete baseline HCSE in both metrics and achieves the best performance in DP across all datasets.
This outcome verifies that the continuous optimization of a relaxed CSE objective in HypCSE is effective for hierarchical clustering.
Regarding the SE metric, HypCSE achieves the best performance on 4 datasets and runner-up performance on 1 dataset.
DPClusterHSBM achieves top-tier SE performance with the lowest SE on Spambase and the second lowest on 4 datasets.
However, the SE metric is heavily dependent on the quality of the constructed similarity graph. 
Consequently, it might not be a reliable indicator when the similarities between data points are not accurately measured.
On the Br. Cancer dataset, although HypCSE achieves the highest DP score, it performs poorly in the SE metric, indicating that the constructed graphs on this dataset fail to fully capture the class-discriminative features.
HypCSE addresses this by learning an improved graph structure via the GSL module.
We also report Dasgupta's costs of trees from all methods in Appendix F.
The ranks of each method in Dasgupta's cost are similar to SE, indicating the consistency between 2 metrics.

\subsection{Further Analysis}
\noindent\textbf{Ablation Study.}
To verify the effectiveness of two key components in the GSL module, we conduct an ablation study and report the experimental results in Table \ref{tab:ablation}.
The base model represents the hyperbolic hierarchical clustering module.
Two key components in the GSL module are the Graph Learning (GL) and the Contrastive Learning (CL) components.
We remove GL by replacing the learner graph $G_l$ with the anchor graph $G_a$ and removing the graph learner $g(\cdot)$.
From Table \ref{tab:ablation}, we find that both GL and CL components improve the overall performance.
Without the guidance of the CL component, the GL component has a minimal effect on its own.
The CL component improves overall performance by learning more discriminative features.


\begin{figure}[t]
\centering
\begin{minipage}{1\linewidth}
    \centering
    \includegraphics[width=\linewidth,trim={0 0 0 0}, clip]{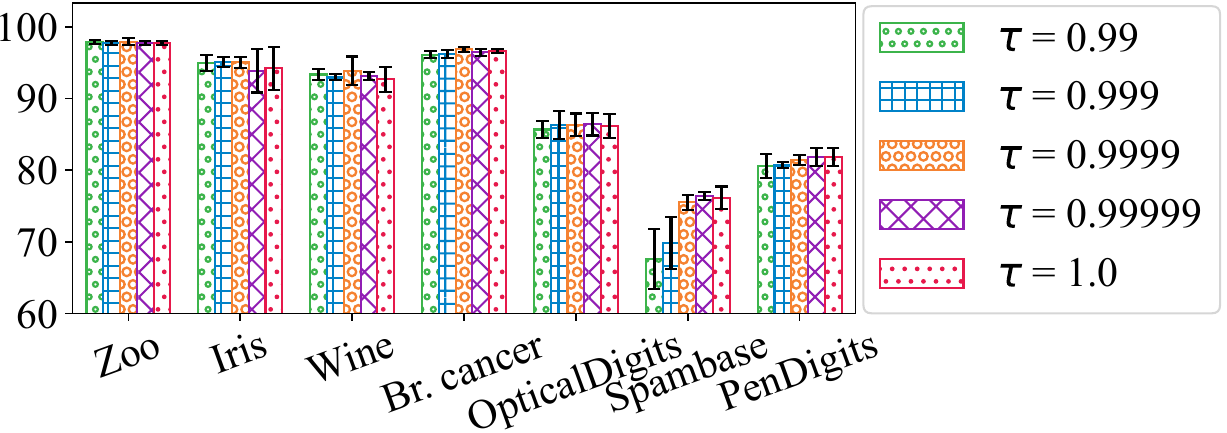}
\end{minipage}
\caption{Parameter sensitivity on decay rate $\tau$ (DP $\%$).
}
\label{fig_tau}
\end{figure}


\begin{figure}[t]
\centering
\begin{minipage}{1.0\linewidth}
    \centering
    \includegraphics[width=\linewidth,trim={0 0 0 0}, clip]{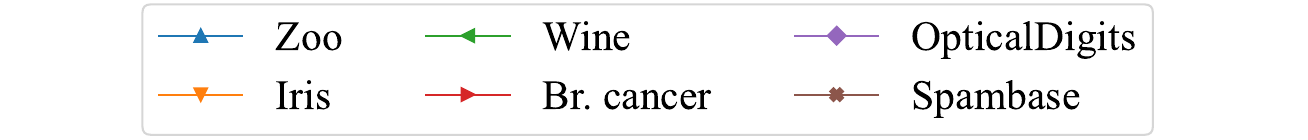}
\end{minipage}
\begin{minipage}{0.49\linewidth}
    \centering
    \includegraphics[width=\linewidth,trim={0 0 0 0}, clip]{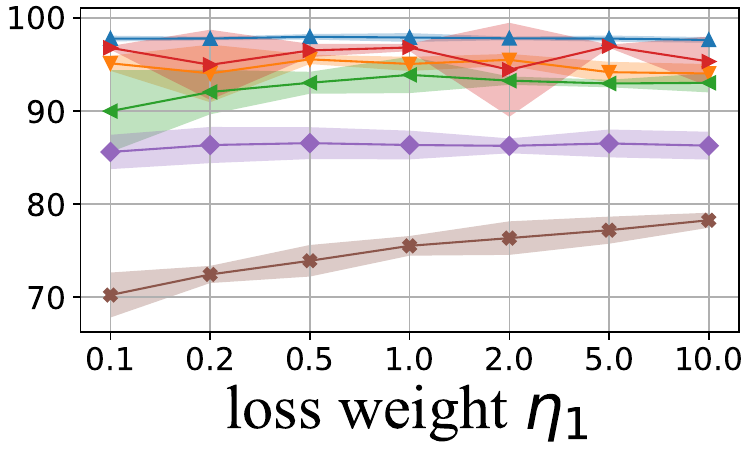}
\end{minipage}
\begin{minipage}{0.49\linewidth}
    \centering
    \includegraphics[width=\linewidth,trim={0 0 0cm 0cm}, clip]{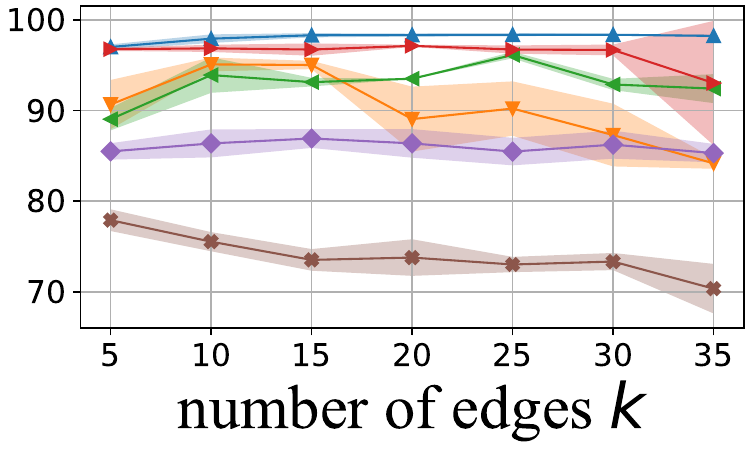}
\end{minipage}
\caption{Parameter sensitivity on $\eta_1$ and $k$ (DP $\%$).
}
\label{fig_sen}
\end{figure}

\noindent\textbf{Parameter Sensitivity.}
We further investigate the parameter sensitivity of HypCSE.
The graph learning decay rate $\tau$ controls the update speed of the anchor graph $G_a$, where a smaller $\tau$ leads to faster updates.
We report the DP scores of HypCSE across different $\tau$ values, as shown in Figure \ref{fig_tau}.
HypCSE showed stable performance across most datasets, with the notable exception of the Spambase dataset. 
This may indicate an ambiguous hierarchical structure within Spambase, which complicates graph structure learning.
While the optimal $\tau$ varies by dataset, we observe that $\tau=0.9999$ yields the best performance across 5 datasets, leading us to adopt it as the default value.
We also investigate the sensitivity of contrastive loss weight $\eta_1$ and the number of edges $k$ when constructing $G_a$, and report the DP scores in Figure \ref{fig_sen}.
We find that $\eta_1$ rarely affects model performance, except for the Spambase dataset.
On Spambase, HypCSE performs better with larger $\eta_1$ values, indicating the significant role of the contrastive loss $\mathcal{L}_{\mathrm{con}}$ for this dataset.
Similarly, the influence of $k$ for the overall performance is also small, except for the Iris and Spambase datasets.
On these datasets, performance drops as more edges are retained, suggesting an increase in noise.
We simply set $\eta_1=1$ and $k=10$ for all datasets.
More parameter sensitivity analysis can be found in Appendix F.2.

\noindent\textbf{Flexibility Analysis.}
Compared to discrete hierarchical clustering methods, continuous methods offer greater flexibility, as they can be combined with other gradient descent optimization-based methods.
Following \cite{chami2020trees}, we evaluate the performance of HypCSE with a classification loss for the similarity-based classification task.
We split 4 datasets from the UCI ML Repository into training, test, and validation sets (30/60/10$\%$ splits), where similarities of all data points are available during training, and the test and validation set labels are only available during testing and validation.
To ensure that only similarities are available for classification, we utilize the similarities of each data point to all other points as the data feature and reduce the feature dimension to 10 via Principal Component Analysis.
We compare HypCSE against HypHC End-to-End learning with a classification loss (HypHC-ETE), HypHC embed-then-classify Two-Step approach (HypHC-TS), and a similarity-based semi-supervised learning method, Label Propagation (LP).
The experimental results are reported in Table \ref{tab:flex}.
Similar to HypHC, the proposed HypCSE can also be jointly optimized with the classification loss.
HypCSE achieves the highest ACC on all datasets, demonstrating its flexibility in jointly integrating with other machine learning pipelines.

\begin{table}[t]
\centering

 {\fontsize{9}{11}\selectfont
 {\setlength{\tabcolsep}{1.5mm}
\begin{tabular}{l|cccc}
\toprule
Methods                    & Zoo & Iris & Glass & Seg.     \\
\midrule
LP                     & 41.4 & 76.7 & 46.8 & 65.3    \\
HypHC-TS        &  84.8$\pm3.5$ & 84.4$\pm1.7$ & 50.6$\pm2.6$ & 64.1$\pm0.9$ \\
HypHC-ETE       &  87.9$\pm3.8$ & 85.6$\pm0.8$ & 54.4$\pm2.9$ & 67.7$\pm3.4$  \\
HypCSE                 &  \textbf{88.3$\pm3.8$} & \textbf{91.8$\pm3.3$} & \textbf{54.7$\pm5.5$} & \textbf{78.8$\pm2.3$}  \\
\bottomrule
\end{tabular}
}
}
\caption{Results (ACC $\%$) of similarity-based classification.}
\label{tab:flex}
\end{table}

\noindent\textbf{Visualization.}
We visualize the partitioning trees generated by HypCSE on a Poincar\'e disk and normalize the tree leaves to lie on the disk border.
The Zoo dataset has 7 classes and the Wine dataset has 3 classes, each with a dominant class (characterized by a large number of data points) in colors blue and orange, respectively.
For both partitioning trees, most leaves from the same class are correctly grouped into the same subtrees.
For Zoo, several leaves of the class in pink are mis-clustered into nearby subtrees.
For Wine, the leaves of the class in orange are too large in number, so some of them are mis-clustered into the other two subtrees.
In all, the visualization of trees verifies the effectiveness of HypCSE in graph embedding and hierarchical clustering.

\begin{figure}[t]
\centering
\begin{minipage}{0.43\linewidth}
    \centering
    \includegraphics[width=\linewidth,trim={1.7cm 1.5cm 1.3cm 1.5cm}, clip]{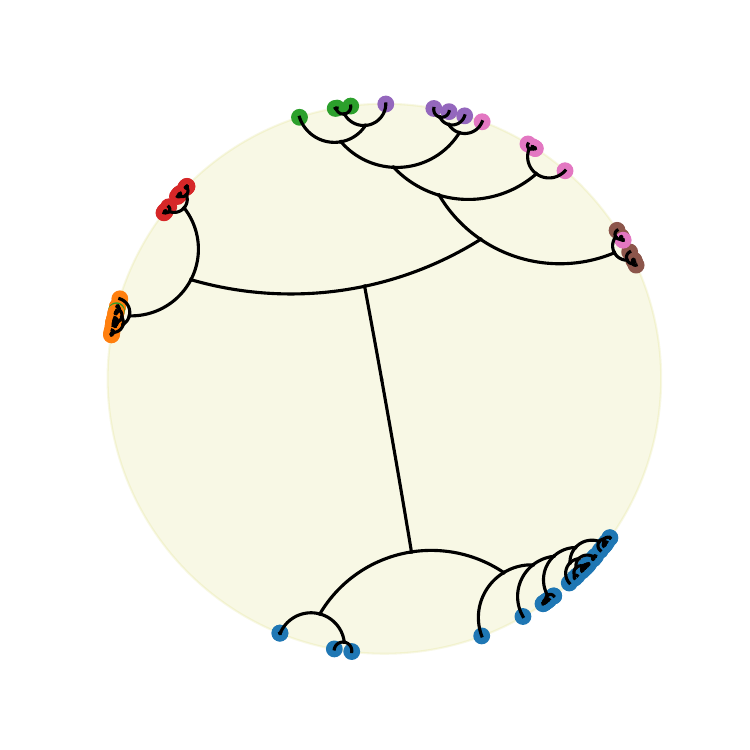}
    \subcaption{Zoo}
    \label{fig_sub_1}
\end{minipage}
\begin{minipage}{0.43\linewidth}
    \centering
    \includegraphics[width=\linewidth,trim={1.7cm 1.5cm 1.3cm 1.5cm}, clip]{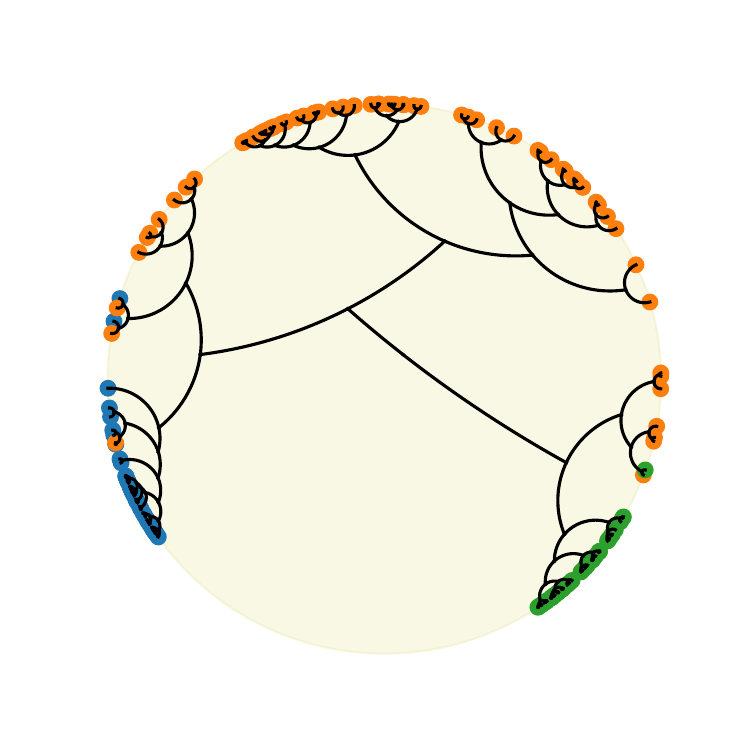}
    \subcaption{Wine}
    \label{fig_sub_2}
\end{minipage}
\caption{Visualization of partitioning trees in space $\mathbb{B}^2$.
}
\label{fig_vis}
\end{figure}

\section{Related Work}
Hierarchical clustering algorithms group data points into nested clusters organized as a dendrogram, and they can be categorized into discrete and continuous optimization methods \cite{zugner2022end}.
Conventional discrete algorithms include bottom-up agglomerative and top-down divisive methods \cite{ran2023comprehensive}.
In contrast, continuous methods learn dendrograms via gradient descent and offer advantages in flexibility, allowing for joint optimization with other pipelines \cite{chami2020trees}.
Before Dasgupta's cost was proposed \cite{dasgupta2016cost}, most existing methods lacked a global objective function \cite{chierchia2019ultrametric}. SE \cite{li2016structural} also serves as a global cost objective for hierarchical clustering \cite{pan2021information} by quantifying information in graphs using partitioning trees.

\section{Conclusion}
In this paper, we introduce HypCSE, a continuous approach for structure-enhanced hierarchical clustering from the perspective of structural entropy.
We develop continuous structural entropy (CSE) by reformulating the classic structural entropy objective using the lowest common ancestor on trees and approximating it in hyperbolic space.
This differentiable formulation enables HypCSE to minimize CSE while simultaneously optimizing other objectives via gradient descent, offering significant flexibility.
To avoid relying on predefined and potentially suboptimal graphs, HypCSE dynamically updates graph structures under the guidance of hyperbolic contrastive learning.
Extensive experiments on 7 real-world datasets demonstrate the superiority of HypCSE.


\setcounter{secnumdepth}{2}

\appendix
\section{Proofs}

\subsection{Proof of Lemma 3}
The inequity in Lemma 3 has been proven for differential structural entropy defined by level-wise assignment matrices \cite{sunlsenet}, and for two-dimensional structural entropy defined on graph partitioning \cite{li2016structural}.
Here, we prove Lemma 3 on high-dimensional structural entropy for partitioning trees.

\noindent\textbf{Lemma 3} (Connection to Graph-Based Clustering).
\emph{
Given an weighted undirected graph $G=(V, E, \mathbf{W})$, $\rho(G) = \mathcal{H}(G) / \mathcal{H}^1(G)$ is the normalized structural entropy, and $\Phi(G) = \min_{S \subseteq V} \frac{\mathrm{cut}(S)}{\min \{ \mathcal{V}_S, \mathcal{V}_{V \backslash S } \}}$ is the graph conductance, the following inequality holds:
\begin{equation}
\rho^\mathcal{T}(G) \geq \Phi(G).
\end{equation}
}

\noindent\emph{Proof.}
For each tree node $\alpha$, its associated vertex set is $T_{\alpha}$ and its cut is $g_\alpha$, its graph conductance is calculated as follows:
\begin{equation}
\phi(T_{\alpha}) = \frac{g_\alpha}{\min \{ \mathcal{V}_{\alpha}, \mathcal{V}_G - \mathcal{V}_\alpha \}}.
\end{equation}
Without loss of generality, we assume that $\mathcal{V}_\alpha \leq \mathcal{V}_G - \mathcal{V}_\alpha$ such that $\min\{ \mathcal{V}_{\alpha}, \mathcal{V}_G - \mathcal{V}_\alpha \} = \mathcal{V}_\alpha$.
The conductance of the graph $G$ is defined as:
\begin{equation}
\Phi(G) = \min_{S \subseteq V} \Phi(S) = \min \{ \frac{\mathrm{cut}(S)}{\min\{ \mathcal{V}_S, \mathcal{V}_{V \backslash S} \} } \},
\end{equation}
where $S$ traverses all possible vertex subsets of $V$.
According to Definition 1, the structural entropy of graph $G$ with a partitioning tree $\mathcal{T}$ can be formulated as follows:
\begin{equation} \label{eq:htg_expand1}
\begin{aligned}
\mathcal{H}^\mathcal{T}(G) &= \sum_{\alpha \in \mathcal{T}, \alpha \neq \lambda} - \frac{g_\alpha}{\mathcal{V}_G} \log_2 \frac{\mathcal{V}_\alpha}{\mathcal{V}_{\alpha^-}} \\
&= - \sum_{\alpha \in \mathcal{T}, \alpha \neq \lambda} \phi(T_\alpha) \frac{\mathcal{V}_\alpha}{\mathcal{V}_G} \log_2 \frac{\mathcal{V}_\alpha}{\mathcal{V}_{\alpha^-}} \\
&\geq - \Phi(G) \sum_{\alpha \in \mathcal{T}, \alpha \neq \lambda} \frac{\mathcal{V}_\alpha}{\mathcal{V}_G} \log_2 \frac{\mathcal{V}_\alpha}{\mathcal{V}_{\alpha^-}}, \\
\end{aligned}
\end{equation}
where $\Phi(G) \leq \phi(T_{\alpha})$ for all tree nodes $\alpha$.

Next, we intend to prove the following:
\begin{equation}\label{eq:additive}
\mathcal{H}^1(G) = - \sum_{\alpha \in \mathcal{T}, \alpha \neq \lambda} \frac{\mathcal{V}_\alpha}{\mathcal{V}_G} \log_2 \frac{\mathcal{V}_\alpha}{\mathcal{V}_{\alpha^-}}.
\end{equation}
Let us subtract the right-hand side from the left-hand side of Equation (\ref{eq:additive}) as follows:
\begin{equation}
\begin{aligned}
A &= \mathcal{H}^1(G) + \sum_{\alpha \in \mathcal{T}, \alpha \neq \lambda} \frac{\mathcal{V}_\alpha}{\mathcal{V}_G} \log_2 \frac{\mathcal{V}_\alpha}{\mathcal{V}_{\alpha^-}} \\
&= - \sum_{i=1}^n \frac{d_i}{\mathcal{V}_G} \log_2 \frac{d_i}{\mathcal{V}_G} + \sum_{\alpha \in \mathcal{T}, \alpha \neq \lambda} \frac{\mathcal{V}_\alpha}{\mathcal{V}_G} \log_2 \frac{\mathcal{V}_\alpha}{\mathcal{V}_{\alpha^-}} \\
&= - \sum_{i=1}^n \frac{d_i}{\mathcal{V}_G} \log_2 \frac{d_i}{\mathcal{V}_G} + \sum_{\alpha \in \mathcal{T}}^{\alpha \neq \lambda} \left[ \frac{\mathcal{V}_\alpha}{\mathcal{V}_G} \log_2 \frac{\mathcal{V}_\alpha}{\mathcal{V}_G} - \frac{\mathcal{V}_\alpha}{\mathcal{V}_G} \log_2 \frac{\mathcal{V}_{\alpha^-}}{\mathcal{V}_G} \right] \\
&= \sum_{\beta \in \mathcal{T}, |T_{\beta}|>1}^{\beta \neq \lambda} \frac{\mathcal{V}_\beta}{\mathcal{V}_G}\log_2 \frac{\mathcal{V}_\beta}{\mathcal{V}_G} - \sum_{\alpha \in \mathcal{T}}^{\alpha \neq \lambda} \frac{\mathcal{V}_\alpha}{\mathcal{V}_G} \log_2 \frac{\mathcal{V}_{\alpha^-}}{\mathcal{V}_G},
\end{aligned}
\end{equation}
where $|T_\beta|=1$ means the associated vertex set of tree $\beta$ has a size of $1$, i.e., $\beta$ is a tree leaf.
Since $\mathcal{V}_\lambda = \mathcal{V}_G$, we have $\frac{\mathcal{V}_\beta}{\mathcal{V}_G} \log_2 \frac{\mathcal{V}_\lambda}{\mathcal{V}_G} = 0$.
Thus, the process continues as follows:
\begin{equation}
\begin{aligned}
A &= \sum_{\beta \in \mathcal{T}}^{|T_{\beta}|>1} \frac{\mathcal{V}_\beta}{\mathcal{V}_G}\log_2 \frac{\mathcal{V}_\beta}{\mathcal{V}_G} - \sum_{\alpha \in \mathcal{T}}^{\alpha \neq \lambda} \frac{\mathcal{V}_\alpha}{\mathcal{V}_G} \log_2 \frac{\mathcal{V}_{\alpha^-}}{\mathcal{V}_G} \\ 
&= \sum_{\beta \in \mathcal{T}}^{|T_{\beta}|>1} \left( \sum_{\gamma \in \mathrm{children}(\beta)}  \frac{\mathcal{V}_\gamma}{\mathcal{V}_G} \right) \log_2 \frac{\mathcal{V}_\beta}{\mathcal{V}_G} - \sum_{\alpha \in \mathcal{T}}^{\alpha \neq \lambda} \frac{\mathcal{V}_\alpha}{\mathcal{V}_G} \log_2 \frac{\mathcal{V}_{\alpha^-}}{\mathcal{V}_G} \\ 
&= \sum_{\gamma \in \mathcal{T}}^{\gamma \neq \lambda}    \frac{\mathcal{V}_\gamma}{\mathcal{V}_G}  \log_2 \frac{\mathcal{V}_{\gamma^-}}{\mathcal{V}_G} - \sum_{\alpha \in \mathcal{T}}^{\alpha \neq \lambda} \frac{\mathcal{V}_\alpha}{\mathcal{V}_G} \log_2 \frac{\mathcal{V}_{\alpha^-}}{\mathcal{V}_G} \\ 
&= 0,
\end{aligned}
\end{equation}
where $\mathrm{children}(\beta)$ is the children set of tree node $\beta$.
Since the equality in Equation (\ref{eq:additive}) holds, we continue Equation (\ref{eq:htg_expand1}) as follows:
\begin{equation}
\begin{aligned}
\mathcal{H}^\mathcal{T}(G) &\geq - \Phi(G) \mathcal{H}^1(G).
\end{aligned}
\end{equation}
Since $\mathcal{H}^\mathcal{T}(G) \geq - \Phi(G) \mathcal{H}^1(G) > 0$, we have:
\begin{equation}
\rho^\mathcal{T}(G) = \frac{\mathcal{H}^\mathcal{T}(G)}{\mathcal{H}^1(G)} \geq \Phi(G).
\end{equation}
Since the above inequality holds for any partitioning tree $\mathcal{T}$ of graph $G$, $\rho^\mathcal{T}(G) \geq \Phi(G)$ holds.

\subsection{Proof of Lemma 5}
\textbf{Lemma 5} (Descendant via LCA). 
\emph{For leaves $\mathcal{T}_i, \mathcal{T}_j$ and $\mathcal{T}_k$ in rooted tree $\mathcal{T}$, $\mathcal{T}_k$ is the descendant of $\mathcal{T}_i \vee \mathcal{T}_j$ if and only if both the following statements hold: 
\begin{equation}\label{eq_lemma5}
    d_\lambda (\mathcal{T}_i \vee \mathcal{T}_j) \leq d_\lambda (\mathcal{T}_i \vee \mathcal{T}_k), \quad d_\lambda (\mathcal{T}_i \vee \mathcal{T}_j) \leq d_\lambda (\mathcal{T}_j \vee \mathcal{T}_k),
\end{equation}
where $d_\lambda (\mathcal{T}_i)$ is the distance between $\mathcal{T}_i$ and the root $\lambda$ in $\mathcal{T}$.}

\noindent\emph{Proof.}
All trees in this paper are unweighted rooted trees.
To prove Lemma 5, we need to prove the following:
\begin{enumerate}
\item If two statements in Equation (\ref{eq_lemma5}) hold, then $\mathcal{T}_k$ is the descendant of $\mathcal{T}_i \vee \mathcal{T}_j$.
\item If $\mathcal{T}_k$ is the descendant of $\mathcal{T}_i \vee \mathcal{T}_j$, then two statements in Equation (\ref{eq_lemma5}) hold.
\end{enumerate}
\noindent\textbf{Subproof 1.} 
Both $\mathcal{T}_i \vee \mathcal{T}_j$ and $\mathcal{T}_i \vee \mathcal{T}_k$ are the ancestors of $\mathcal{T}_i$. 
Since each tree node has only one ancestor, the above tree nodes are in an ancestor-descendant chain, which means $\mathcal{T}_i \vee \mathcal{T}_j$ is the ancestor, descendant, or $\mathcal{T}_i \vee \mathcal{T}_k$ itself. 
Given that $d_\lambda (\mathcal{T}_i \vee \mathcal{T}_j) \leq d_\lambda (\mathcal{T}_i \vee \mathcal{T}_k)$, we have $\mathcal{T}_i \vee \mathcal{T}_j$ is the ancestor of $\mathcal{T}_i \vee \mathcal{T}_k$, or $\mathcal{T}_i \vee \mathcal{T}_j$ is $\mathcal{T}_i \vee \mathcal{T}_k$. Since $\mathcal{T}_k$ is the descendant of $\mathcal{T}_i \vee \mathcal{T}_k$, we have $\mathcal{T}_k$ is the descendant of $\mathcal{T}_i \vee \mathcal{T}_j$. 
This completes \textbf{Subproof 1}.

\noindent\textbf{Subproof 2.} 
Given that $\mathcal{T}_k$ is the descendant of $\mathcal{T}_i \vee \mathcal{T}_j$, since $\mathcal{T}_i$ is the descendant of $\mathcal{T}_i \vee \mathcal{T}_j$, $\mathcal{T}_i \vee \mathcal{T}_j$ is a Common Ancestor of $\mathcal{T}_i$ and $\mathcal{T}_k$.
In that case, the LCA $\mathcal{T}_i \vee \mathcal{T}_k$ is the descendant of $\mathcal{T}_i \vee \mathcal{T}_j$, or $\mathcal{T}_i \vee \mathcal{T}_k$ is $\mathcal{T}_i \vee \mathcal{T}_j$.
Thus, $d_\lambda (\mathcal{T}_i \vee \mathcal{T}_j) \leq d_\lambda (\mathcal{T}_i \vee \mathcal{T}_k)$ holds.
For the same reason, $d_\lambda (\mathcal{T}_i \vee \mathcal{T}_j) \leq d_\lambda (\mathcal{T}_j \vee \mathcal{T}_k)$ holds.
This completes \textbf{Subproof 2}.

\section{Hyperbolic Tree Decoding Algorithm}
To help tree leaves separate in the Poincar\'e disk, we normalize their embeddings to lie on the hyperbolic diameter \cite{chami2020trees}.
We merge clusters until all clusters are merged into one binary partitioning tree.
This tree decoding procedure is summarized in Algorithm \ref{algo:dec}.

\begin{algorithm}[t]
\caption{Tree decoding given embedding $\mathbf{Z}_{\mathbb{B}}$}
\label{algo:dec}
\begin{algorithmic}[1]
    \STATE \textbf{Input:} Embeddings of vertices $\mathbf{Z}_{\mathbb{B}} = \{\mathbf{z}_1,...,\mathbf{z}_n\}$;
    \STATE \textbf{Output:} Binary partitioning tree $\mathcal{T}$.
    \STATE Initialize $\mathcal{T}$ to contain leaves $\{\mathbf{z}_1,...,\mathbf{z}_n\}$ and root $\lambda$
    \STATE Set each data point $\mathbf{z}_i$ as a separate cluster $C_i = \{ \mathbf{z}_i \}$
    \STATE Calculate the closeness of all cluster pairs
    \WHILE{number of clusters $>2$}
        \STATE Merge $(C_\alpha,C_\beta)$ into a new cluster $C_\gamma = C_\alpha \cup C_\beta$ conditioned on $\arg \min_{\alpha, \beta} \{ \mathrm{closeness}(C_\alpha,C_\beta) \}$
        \FOR{each remaining cluster $C_\delta$}
            \STATE Calculate the closeness between $C_\gamma$ and $C_\delta$
        \ENDFOR
    \ENDWHILE
    \STATE Set the remaining two clusters as the children of $\lambda$
    \RETURN $\mathcal{T}$ (which is a binary tree)
\end{algorithmic}
\end{algorithm}

\section{Time Complexity \& Scalability}
Without tailored scalability strategies, the overall time complexity of HypCSE is $O(n^2 \times n_{epoch})$, where $n$ is the number of data points and $n_{epoch}$ is the number of epochs in the training procedure.
To facilitate hierarchical clustering on larger datasets, we apply several scalability strategies to improve the efficiency of HypCSE.
The resulting time complexity is then $O(n\times n' \times n_{epoch} + n \times \log n \times n_{epoch})$, where $n'$ is the subgraph sampling size.
Since both the sampling size $n'$ and the number of epochs $n_{epoch}$ are constants, the time complexity of HypCSE can be given as $O(n \log n)$.

\subsection{Time Complexity}
HypCSE consists of two major modules, the hyperbolic hierarchical clustering module and the graph structure learning (GSL) module.
In the hyperbolic hierarchical clustering module, the graph construction step consists of finding $k$ nearest neighbors of data points, which requires time of $O(k \times n\log n)$.
The hyperbolic graph embedding step requires time of $O(|E|)$ each epoch.
The CSE loss calculation step consists of calculating the hyperbolic distance of the geodesic to the origin $d_o^{\mathbb{B}}(\mathbf{z}_i \vee \mathbf{z}_j)$ for any data point pairs and then calculating the loss $\mathcal{L}_{\mathrm{cse}}$ in Eq. (8), which requires time of $O(n^2)$ and $O(|E|)$ each epoch, respectively.
The tree decoding step consists of calculating and sorting the closeness of all vertex pairs, which requires time of $O(n^2)$.
In the GSL module, the vertex embeddings calculation in Eq. (13) requires time of $O(|E|)$,
the graph construction step requires time of $O(n\log n)$, and the graph updating step requires time of $O(E)$ each epoch.
Taken together, the time complexity of HypCSE is $O(n^2 \times n_{epoch})$.

\subsection{Scalability}
Scalability is a crucial factor for hierarchical clustering algorithms.
We adopt three strategies to improve the scalability of HypCSE, including subgraph sampling to reduce the graph size, fast tree decoding to avoid quadratic decoding time, and sparse graph updating to avoid dense graphs.

\noindent\textbf{Subgraph Sampling.}
Without scalability strategies, the time complexity of HypCSE is $O(n^2 \times n_{epoch})$, which largely depends on the number of data points $|V| = n$.
We propose sampling subgraphs of size $n'$ for minibatch training, which significantly reduces the time complexity of HypCSE.
After constructing graph $G$ from the dataset, we sample subgraphs of size $n'$ in each epoch and train one subgraph in each minibatch.
To avoid performance degradation by subgraph sampling \cite{chiang2019cluster}, we adopt a neighborhood-preserving graph sampling strategy \cite{zeng2024scalable} to retain as much as edges as possible.
Specifically, given a graph $G$, we sample a subgraph by first establishing an initial selected vertex set $S_{select}$, where we randomly choose $n_{seed}$ vertices as sampling seeds.
Next, for each vertex currently in $S_{select}$, we iteratively extend the set by adding its neighbor vertices in $G$.
This iterative expansion continues until the size of $S_{select}$ reaches a predefined sampling size threshold $n'$.
We sample a list of unique vertex sets from $G$ without replacement, each of which corresponds to a subgraph for minibatch training.
In each epoch, graph $G$ is sampled into $\lfloor n / n' \rfloor$ subgraphs of size $n'$, each containing $|E| / n'$ edges.
The time complexity of CSE loss calculation step is then reduced to $O(\lfloor n / n' \rfloor \times ( n'^2 + |E|/n'))$ = $O(n \times n' + |E|)$.

\noindent\textbf{Fast Tree Decoding.}
Tree decoding in Algorithm 1 involves calculating the closeness of all vertex pairs, which requires a time complexity of $O(n^2)$.
Moreover, the following bottom-up clustering process, using optimized algorithms like SLINK \cite{sibson1973slink}, requires a time complexity of $O(n^2)$.
To enhance the scalability of tree decoding in HypCSE, we adopt sparse graph structures by the K-Nearest-Neighbor algorithm and the minimum spanning tree trick.
Specifically, after obtaining the vertex embeddings $\mathbf{Z}_{\mathbb{B}} = \{ \mathbf{z}_1, \ldots, \mathbf{z}_n \}$, we construct a sparse graph $G_\mathbb{B}$ by K-Nearest-Neighbor algorithm \cite{pedregosa2011scikit}, which requires time of $O(n \log n)$.
Since the minimum spanning tree algorithm is equivalent to the single-linkage algorithm \cite{gower1969minimum}, we build the binary partitioning tree using Kruskal's algorithm \cite{kruskal1956shortest}. 
This can be completed in time of $O(|E_{\mathbb{B}}| \log |E_{\mathbb{B}}|)$, where $|E_{\mathbb{B}}|$ is the number of edges in $G_\mathbb{B}$.
Since each vertex in $G_\mathbb{B}$ connect to only K edges, we have $O(|E_{\mathbb{B}}| \log |E_{\mathbb{B}}|) = O(n \log n)$.
Taken together, tree decoding with sparse graphs has a complexity of $O(n \log n)$.

\noindent\textbf{Sparse Graph Updating.}
In the GSL module, for a given anchor graph $G_a$ with the affinity matrix $\mathbf{A}_a$, we learn a learner graph $G_l$ with the affinity matrix $\mathbf{A}_l$.
Simply updating the anchor graph using affinity matrices \cite{liu2022towards} can lead to dense graphs: 
\begin{equation}
\mathbf{A}_a \leftarrow \tau \mathbf{A}_a + (1-\tau) \mathbf{A}_l,
\end{equation}
which requires time of $O(n^2)$
Moreover, since the edges in $\mathbf{A}_l$ are different from $\mathbf{A}_a$, the anchor graph $G_a$ becomes a dense graph after several epochs.
This dense graph will largely decrease the efficiency of HypCSE.
Instead, we update the anchor graph by edge sampling, as in Eq. (13), which requires a time complexity of $O(|E|)$ and avoids maintaining a dense anchor graph.

With the above scalability strategies, the time complexity of HypCSE is then $O(n \times n' \times n_{epoch} + n \times \log n \times n_{epoch})$.

\section{Further Related Works.}

\noindent\textbf{Structural Information Theory.}
Shannon entropy \cite{shannon1948mathematical} is the foundation of information theory as it quantifies the amount of information of a given variable.
Properly quantifying information for graphs is challenging and important \cite{brooks2003three}.
Unlike early practices defined by unstructured probability distributions \cite{braunstein2006laplacian,bianconi2009entropy}, SE \cite{li2016structural} measures graph complexity by characterizing the uncertainty of the hierarchical topology of graphs.
SE has proven effective in various fields, including bioinformatics \cite{chen2023incorporating}, graph classification \cite{wu2022structural}, and reinforcement learning \cite{zeng2023hierarchical}.
Specifically related to this work, SE has been applied to community detection \cite{zhu2023low}, community deception \cite{liu2019rem}, deep graph clustering \cite{sunlsenet}, and agglomerative hierarchical clustering \cite{pan2021information}.
However, it has not yet been introduced to continuous hierarchical clustering due to the challenge of relaxing the discrete formulation of SE at very high dimensions.

\noindent\textbf{Hierarchical Clustering.}
Hierarchical clustering algorithms partition data points into nested clusters organized as a dendrogram; they can be grouped into discrete and continuous optimization methods \cite{zugner2022end}.
Conventional discrete algorithms include agglomerative methods and divisive methods \cite{ran2023comprehensive}.
For agglomerative methods \cite{ward1963hierarchical}, they set each data point as a cluster and iteratively merge similar clusters into larger clusters.
The key issue is to decide which two clusters to merge in the next round.
For divisive methods \cite{zhao2002evaluation}, they set all data points in a single cluster and iteratively divide the clusters into smaller ones.
The key issue is to decide which cluster to split and how to split the clusters.
In terms of clustering objectives, earlier practices chose local heuristics that can not evaluate the overall performance of the generated dendrograms.
In 2016, Dasgupta proposed a global cost function \cite{dasgupta2016cost} based on LCA for similarity-based hierarchical clustering and introduced a provably good approximation divisive algorithm for optimization.
From then on, other global objectives \cite{charpentier2019tree} including variants of Dasgupta's cost \cite{cohen2019hierarchical,wang2020improved} are proposed, along with various optimization algorithms.
SE quantifies the amount of information in graphs with hierarchical partitioning trees, making it another global cost objective for hierarchical clustering \cite{pan2021information} from the information perspective.

Continuous methods \cite{monath2017gradient,monath2019gradient,chierchia2019ultrametric,chami2020trees,zugner2022end} relax certain global objectives and optimize them using gradient-based optimizers.
They have advantages in terms of flexibility compared to discrete methods, since they can be integrated into commonly used end-to-end learning pipelines.
More recently, hierarchical clustering methods based on continuous optimization have been proposed for scalability \cite{long2023cross} and multiview data \cite{LinBB0ZX22,lin2023mhcn}.
Existing methods overlook the information in the data features.
Our work HypCSE learns better graph structures by a GCN-based graph learner and contrastive learning and learns graph embeddings in hyperbolic space for binary hierarchical clustering.

\noindent\textbf{Hierarchy Learning in Hyperbolic Space.}
Hyperbolic embeddings have been drawing attention due to their advantage in modeling data with hierarchical structures \cite{peng2021hyperbolic}, such as scRNAseq data \cite{klimovskaia2020poincare,ding2021deep} and taxonomies \cite{nickel2017poincare,nickel2018learning}.
Nickel \textit{et al.} proposed to learn hierarchical representations of taxonomy data via Poincar\'e embeddings based on Riemannian optimization \cite{nickel2017poincare}.
They also show that the Lorentz model is more efficient and stable in learning embeddings and generates embeddings at higher quality compared to the Poincar\'e model \cite{nickel2018learning}.
Later on, various hyperbolic deep neural networks \cite{chami2019hyperbolic,liu2019hyperbolic} analogous to models in Euclidean space have been proposed.
Chen \textit{et al.} propose Fully Hyperbolic Neural Networks \cite{chen2021fully} with both Lorentz boost and rotation and demonstrate their higher representation capability.
Clustering is also an important application of hyperbolic embedding.
LSEnet \cite{sunlsenet} embeds attributed graphs by the Lorentz model for deep graph clustering by minimizing structural entropy.
gHHC \cite{monath2019gradient} and HypHC \cite{chami2020trees} embed vector data by the Poincar\'e model for hierarchical clustering by minimizing Dasgupta's cost.

\section{Detailed Experimental Setup}

\subsection{Datasets}
We evaluate HypCSE on 7 public clustering datasets, including Zoo, Iris, Glass, Segmentation, Wine, Australian, and Breast Cancer, all taken from the UCI Machine Learning Repository \cite{UCI}.
Detailed statistics of these datasets are summarized in Table \ref{tab:datasets}.

\setcounter{table}{3}
\begin{table}[t]
\centering
\caption{Detailed statistics of datasets.}
\label{tab:datasets}
\begin{tabular}{l|ccc}
\toprule
Datasets & \verb|#|Data points & \verb|#|Features & \verb|#|Classes \\ 
\midrule
Zoo & 101 & 16 & 7 \\
Iris & 150 & 4 & 3 \\
Wine & 175 & 13 & 3 \\
Breast Cancer & 683 & 10 & 2 \\
OpticalDigits & 1,797 & 64 & 10 \\
Spambase & 4,601 & 57 & 2 \\
PenDigits & 10,992 & 16 & 10 \\
\bottomrule
\end{tabular}
\end{table}

\begin{table*}[t]
  \centering

    \caption{Hierarchical clustering quality measured in Dasgupta's cost. \textbf{Bold}: the best performance, \underline{underline}: the runner-up performance.}
    \label{table_da}
  \begin{tabular}{l|ccccccc}
    \toprule
    Methods
    & Zoo & Iris & Wine & Br. Cancer & OpticalDigits & Spambase & PenDigits
    \\
    \midrule 
    SingleLinkage & 4499.992 & 45380.796 & 67920.236 & 1224116.656 & 345332.736 & 70302109.781 & 76567530.892 \\
    BKM &     7849.072 & 54788.642 & 54693.242 & 693561.766 & 745491.549 & 41204923.569 & \underline{40284225.153} \\
    HDBSCAN & 5571.228 & 60452.402 & 95907.829 & 1491110.118 & 748313.231 & 80716545.145 & 132350602.149 \\
    HCSE & \underline{4175.671} & 41380.998 & 58347.563 & 592273.513 & 310539.268 & 45263227.036 & 59508491.488 \\
    SpecWRSC & 6144.790 & 35099.430 & \underline{47293.848} & \underline{537871.384} & \underline{257737.166} & 101118252.200 & 92647589.507 \\
    DPClusterHSBM & 6524.713 & \underline{30035.403} & 51149.127 & 660334.272 & 287088.373 & \textbf{34261369.825} & not converge \\
    \midrule
    UFit &  7901.511 & 35457.013 & 56666.424 & 714959.470 & 451331.653 & 59850887.347 & 69377079.903 \\
    HypHC & 4499.059 & 41591.450 & 50855.890 & 914908.598 & 1271341.559 & 79220674.974 & 676405500.194 \\
    FPH   & 4264.423 & 31349.853 & \textbf{43690.747} & \textbf{463823.612} & 321385.930 & \underline{40420419.233} & 47413933.365 \\
    HypCSE & \textbf{3897.084} & \textbf{29281.880} & 47499.454 & 615481.751 & \textbf{184393.801} & 48662062.993 & \textbf{28164085.133} \\
    \bottomrule
    \end{tabular}
\end{table*}

\setcounter{figure}{5}
\begin{figure*}[t]
\centering
\begin{minipage}{0.9\linewidth}
    \centering
    \includegraphics[width=\linewidth,trim={0 0 0 0}, clip]{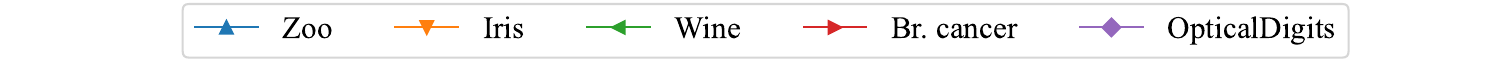}
\end{minipage}
\begin{minipage}{0.245\linewidth}
    \centering
    \includegraphics[width=\linewidth,trim={0 0 0 0}, clip]{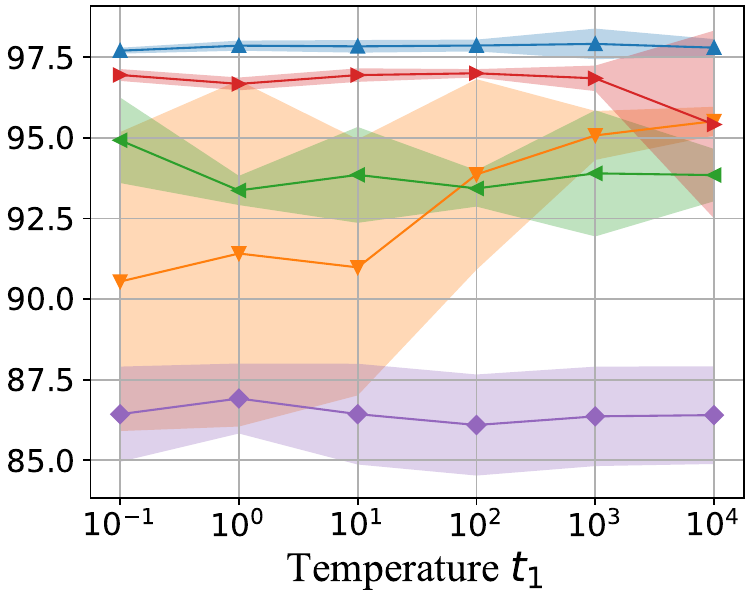}
\end{minipage}
\begin{minipage}{0.245\linewidth}
    \centering
    \includegraphics[width=\linewidth,trim={0 0 0cm 0cm}, clip]{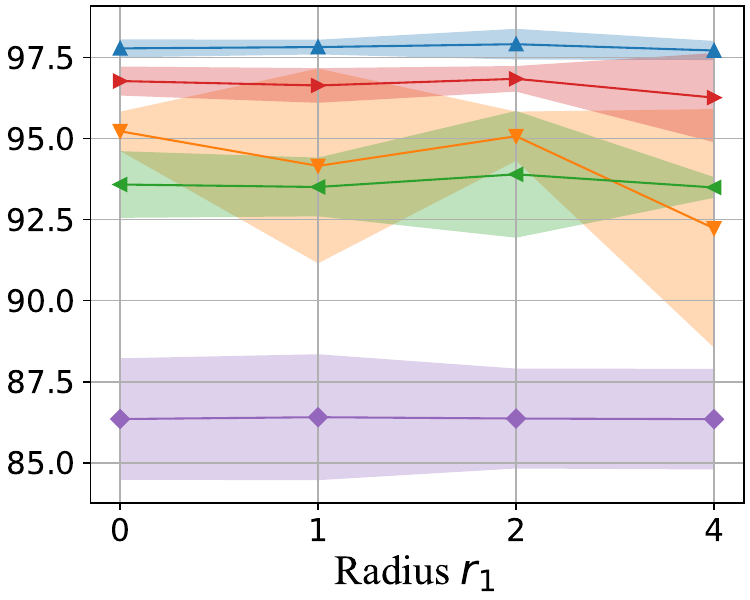}
\end{minipage}
\begin{minipage}{0.245\linewidth}
    \centering
    \includegraphics[width=\linewidth,trim={0 0 0 0}, clip]{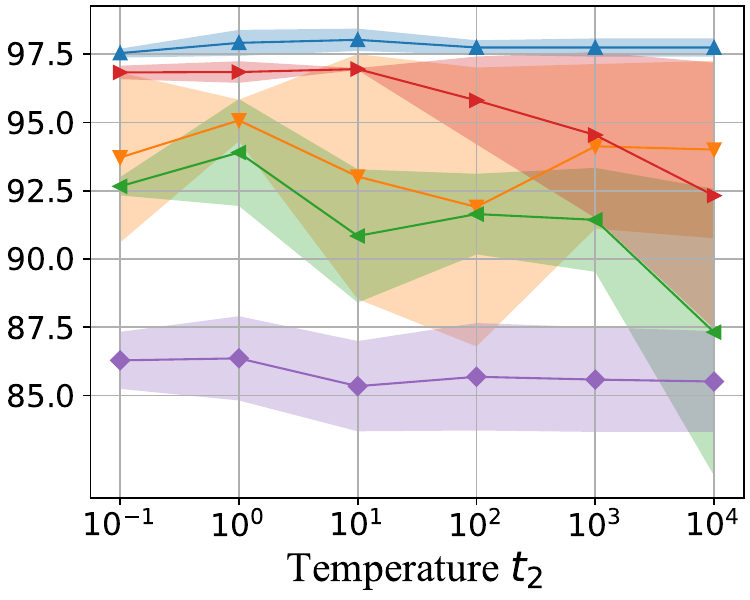}
\end{minipage}
\begin{minipage}{0.245\linewidth}
    \centering
    \includegraphics[width=\linewidth,trim={0 0 0cm 0cm}, clip]{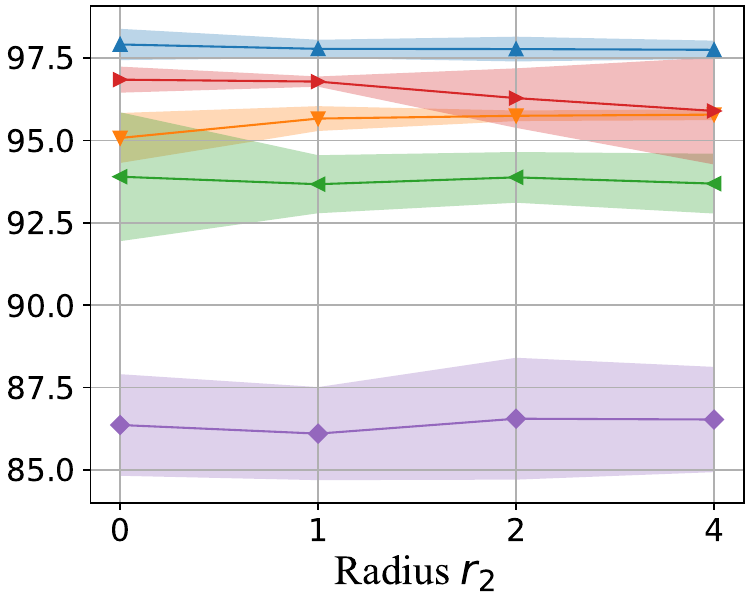}
\end{minipage}
\caption{Hyperparameter sensitivity on $t_1$, $r_1$ in Eq. (9) and $t_2$, $r_2$ in Eq. (15) measured by DP (\%).
}
\label{fig_sentr}
\end{figure*}

\subsection{Baselines}
We compare HypCSE against 9 hierarchical clustering methods, including 6 discrete and 3 continuous methods.
The following is a brief introduction to the methods.

$\bullet$ SpecWRSC \cite{laenen2023nearly} is an efficient top-down discrete hierarchical clustering algorithm based on spectral clustering and the vertex-weighted recursive sparsest cut algorithm.

$\bullet$ DPClusterHSBM \cite{imola2023differentially} is an edge-level differential private discrete hierarchical clustering algorithm based on the hierarchical Stochastic block models.

$\bullet$ FPH \cite{zugner2022end} is a probabilistic model over hierarchies on graphs by continuous relaxation of Dasgupta cost or Tree-Sampling Divergence for continuous hierarchical clustering.

$\bullet$ HCSE \cite{pan2021information} is a discrete hierarchical clustering method via heuristic SE minimization that produces binary partitioning trees by the Stretching operator and converts them to certain tree heights by the Compressing operator.
We adopt binary trees using the Stretching operator for evaluation, as they have higher DPs and lower SEs.

$\bullet$ HypHC \cite{chami2020trees} is a continuous similarity-based hierarchical clustering method that learns hyperbolic embeddings of tree leaves and maps them back into dendrograms for hierarchical clustering.
We adopt the greedy top-down decoding method to obtain the partitioning tree for evaluation, as recommended in the original paper.

$\bullet$ UFit \cite{chierchia2019ultrametric} is a continuous hierarchical clustering algorithm by fitting an ultrametric distance to a dissimilarity graph through optimizing the sum of squared errors between the sought ultrametric and the edge weights of the given graph, together with cluster size regularization.

$\bullet$ HDBSCAN \cite{mcinnes2017hdbscan} is a discrete hierarchical density-based spatial clustering algorithm that performs DBSCAN over varying epsilon values.

$\bullet$ BKM \cite{moseley2017approximation} is a discrete similarity-based analogy of the Hierarchical KMeans method, where Hierarchical KMeans is a discrete top-down hierarchical clustering method based on KMeans algorithm. 

$\bullet$ SingleLinkage \cite{gower1969minimum} is a discrete agglomerative hierarchical clustering that merges clusters containing the closest pair of data points.

\subsection{Evaluation Metrics.}
We adopt a widely used hierarchical clustering metric, Dendrogram Purity (DP) \cite{yadav2019supervised,heller2005bayesian,chami2020trees}, for performance evaluation.
DP is a holistic measure of a cluster tree, defined as the average purity score of the ancestors of all leaf pairs with the same ground-truth labels.
Given hierarchical clustering $\mathcal{T}$ of $\mathbf{X} = \{ \mathbf{x}_1, \mathbf{x}_2,...,\mathbf{x}_n \}$, with ground truth $\mathbf{y}=\{y_1,y_2,...,y_n\}$, the DP of $\mathcal{T}$ is defined as follows:
\begin{equation}
    \mathrm{DP}(\mathcal{T}) = \frac{1}{|\mathcal{W}^\star|} \sum_{\mathbf{x}_i,\mathbf{x}_j \in \mathcal{W}^\star} \mathrm{pur}(\mathrm{lvs}(\mathrm{LCA}(\mathbf{x}_i,\mathbf{x}_j)), \mathcal{C}^\star (y_i)),
\end{equation}
where $\mathcal{W}^\star$ is the set of unordered pairs of data points belonging to the same ground truth class, $\mathcal{C}^\star (y_i)$ is the ground truth class with label $y_i$, $\mathrm{LCA}$ and $\mathrm{lvs}$ represents the lowest common ancestor and leaves of nodes in $\mathcal{T}$, respectively, and $\mathrm{pur}$ represents purity score defined as $\mathrm{pur}(A,B) = |A \cap B| / |A|$.

\subsection{Implementation Details.}
We implement HypCSE using PyTorch 2.0, Geoopt, PyG, and NetworkX.
We train HypCSE for 200 epochs and evaluate the model every epoch by decoding the partitioning tree.
The model that produces the partitioning tree with the lowest SE is selected as the best, and the DP value is calculated based on it.
In the graph structure learning module, we first generate vertex embeddings by a graph learner $g(\cdot)$ and then construct the learner graph.
For 4 small datasets, a two-layer GCN \cite{kipf2017semi} graph learner and cosine similarity are applied.
For 3 larger datasets, a two-layer MLP graph learner and Gaussian similarity with a kernel width of 1 are used instead.
In all visualization cases, we train the model with embedding dimension in the hyperbolic encoder $f(\cdot)$ set to 2 to facilitate the visualization of partitioning trees in Poincar\'e space.
Experiments are conducted on a Linux server with two Intel(R) Xeon(R) Platinum 8336C CPUs, two NVIDIA A800 GPUs, and 500 GB of RAM.

\section{Additional Results}

\subsection{Additional Hierarchical Clustering Quality}
Dasgupta's cost \cite{dasgupta2016cost} is a global cost for evaluating hierarchical clustering partitioning trees, which has been widely adopted as the evaluation metric for hierarchical clustering \cite{chami2020trees,laenen2023nearly,zugner2022end,imola2023differentially}.
A lower Dasgupta's cost means better hierarchical clustering performance.
In Table \ref{table_da}, we also report Dasgupta's costs of trees in the constructed anchor graphs, without updating, from all methods.
We find that the ranks of each method in Dasgupta's cost are similar to SE, indicating that Dasgupta's cost is consistent with SE.
Specifically, HypCSE achieves the best performance on 4 datasets, demonstrating that HypCSE achieves top-tier hierarchical clustering quality in Dasgupta's cost.
SpecWRSC achieves runner-up performance on 3 datasets, but exhibits poor performance on the Spambase and PenDigits datasets.
On Br. Cancer, HypCSE performs poorly in Dasgupta's cost but has the highest DP score, indicating that the constructed graphs on this dataset fail to fully capture the class-discriminative features.
In all, HypCSE is a top-tier hierarchical clustering method when measured in Dasgupta's cost.

\subsection{Hyperparameter Sensitivity}
The impact of different temperature ($t_1, t_2$) and radius ($r_1, r_2$) hyperparameters on HypCSE's performance, as defined in Equation (9) and Equation (15) respectively, is illustrated in Figure \ref{fig_sentr}.
The temperature hyperparameter $t_1$ in Equation (9) influences the continuity of the $\mathcal{L}_{\mathrm{cse}}$ loss. Smaller $t_1$ values provide a better approximation but introduce greater optimization challenges. 
While performance remained stable across datasets, the optimal $t_1$ varied. 
For the Iris dataset, performance improved with larger $t_1$ values, leading us to hypothesize that optimization is particularly difficult on this dataset when $t_1$ is small. We set $ t_1 = 1000$ as the default.
The radius hyperparameter $r_1$ in Equation (9) controls how the hyperbolic LCA distance is converted into a similarity measure. 
This parameter also showed stable performance across datasets, and we've set its default value to $r_1=2$.
Another temperature hyperparameter, $t_2$ in Equation (15), governs the sharpness of the softmax function. 
A low $t_2$ emphasizes hard sample pairs, while a high $t_2$ leads to smoother embeddings.
For the Wine and Breast Cancer datasets, higher $t_2$ values resulted in decreased performance, suggesting that HypCSE struggles with discriminating some hard sample pairs in these cases. 
We've set the default $t_2$ to $1$.
Finally, the radius hyperparameter $r_2$ in Equation (15) also transforms the hyperbolic LCA distance into a similarity measure. 
Performance proved stable across different $r_2$ values, and we establish $r_2=0$ as its default setting.




\section{Acknowledgments}
The corresponding author is Yicheng Pan. 
This work is partly supported by National Key R\&D Program of China (2021YFB3500700), the NSFC through grants 62322202, 62441612, and 62202164, CCF-DiDi GAIA collaborative Research Funds for Young Scholars through grant 202527, National Key Laboratory under grant 241-HF-D07-01, the Fundamental Research Funds for the Central Universities, and State Key Laboratory of Complex \& Critical Software Environment (CCSE-2024ZX-20).

\bibliography{references}


\end{document}